\def\paperID{0000}
\def\confName{CVPR}
\def\confYear{2024}

\def\paperTitle{EG-HumanNeRF: Efficient Generalizable Human NeRF \\ Utilizing Human Prior for Sparse View}

\def\authorBlock{
    Zhaorong Wang
    \qquad
     Yoshihiro Kanamori 
     \qquad
    Yuki Endo \\
    University of Tsukuba \\
    {\tt\small zhaorong.wang1997@gmail.com, \{kanamori, endo\}@cs.tsukuba.ac.jp}
}

\newif\ifreview 
\newif\ifarxiv \newcommand{\arxiv}{\arxivtrue}
\newif\ifcamera 
\newif\ifrebuttal 

\arxiv 

\pdfoutput=1
\documentclass[10pt,twocolumn,letterpaper]{article}
\ifreview \usepackage[review]{cvpr} \fi
\ifarxiv \usepackage[pagenumbers]{cvpr} \fi
\ifrebuttal \usepackage[rebuttal]{cvpr} \fi
\ifcamera \usepackage{cvpr} \fi

\usepackage{graphicx}	
\usepackage{amsmath}	
\usepackage{amssymb}	
\usepackage{booktabs}
\usepackage{times}
\usepackage{microtype}
\usepackage{epsfig}
\usepackage[table,xcdraw,dvipsnames]{xcolor}
\usepackage{caption}
\usepackage{float}
\usepackage{placeins}
\usepackage{color, colortbl}
\usepackage{stfloats}
\usepackage{enumitem}
\usepackage{tabularx}
\usepackage{xstring}
\usepackage{multirow}
\usepackage{xspace}
\usepackage{url}
\usepackage{subcaption}
\usepackage{xcolor}
\usepackage[hang,flushmargin]{footmisc}

\ifcamera \usepackage[accsupp]{axessibility} \fi

\usepackage{comment}
\usepackage{booktabs}
\usepackage{multirow}
\usepackage{graphicx}
\usepackage{xfrac}
\usepackage{placeins}
\usepackage[T1]{fontenc}
\usepackage{lmodern}
\usepackage{type1cm}
\usepackage{diagbox}
\usepackage{pifont}

% ...

\ifarxiv  \fi

\newcommand{\R}[1]{{%
    \textbf{%
        \ifstrequal{#1}{1}{\textcolor{red}{R#1}}{%
        \ifstrequal{#1}{2}{\textcolor{blue}{R#1}}{%
        \ifstrequal{#1}{3}{\textcolor{magenta}{R#1}}{%
        \ifstrequal{#1}{4}{\textcolor{teal}{R#1}}{%
                           \textcolor{cyan}{R#1}%
        }}}}%
    }%
}}

\DeclareMathOperator*{\argmin}{arg\,min}

\newcommand{\cmark}{\ding{51}}%
\newcommand{\xmark}{\ding{55}}%

\newcommand{\knmrA}[1]{#1}
\newcommand{\wA}[1]{#1}

\usepackage{xr-hyper}

\makeatletter
\newcommand*{\addFileDependency}[1]{
  \typeout{(#1)}
  \@addtofilelist{#1}
  \IfFileExists{#1}{}{\typeout{No file #1.}}
}

\makeatother
\newcommand*{\myexternaldocument}[1]{
    \externaldocument{#1}
    \addFileDependency{#1.tex}
    \addFileDependency{#1.aux}
}

\definecolor{cvprblue}{rgb}{0.21,0.49,0.74}
\usepackage[pagebackref,breaklinks,colorlinks,citecolor=cvprblue]{hyperref}
\usepackage[capitalize]{cleveref}
\crefname{section}{Sec.}{Secs.}
\crefname{table}{Table}{Tables}
\crefname{figure}{Fig.}{Figs.}

\ifarxiv \crefname{appendix}{App.}{Apps.}
\else \crefname{appendix}{Suppl.}{Suppls.} \fi

\unless\ifarxiv \myexternaldocument{_supplementary} \fi

\begin{document}
\title{\paperTitle}
\author{\authorBlock}
\twocolumn[{%
\renewcommand\twocolumn[1][]{#1}%
\maketitle
\includegraphics[width=1\linewidth]{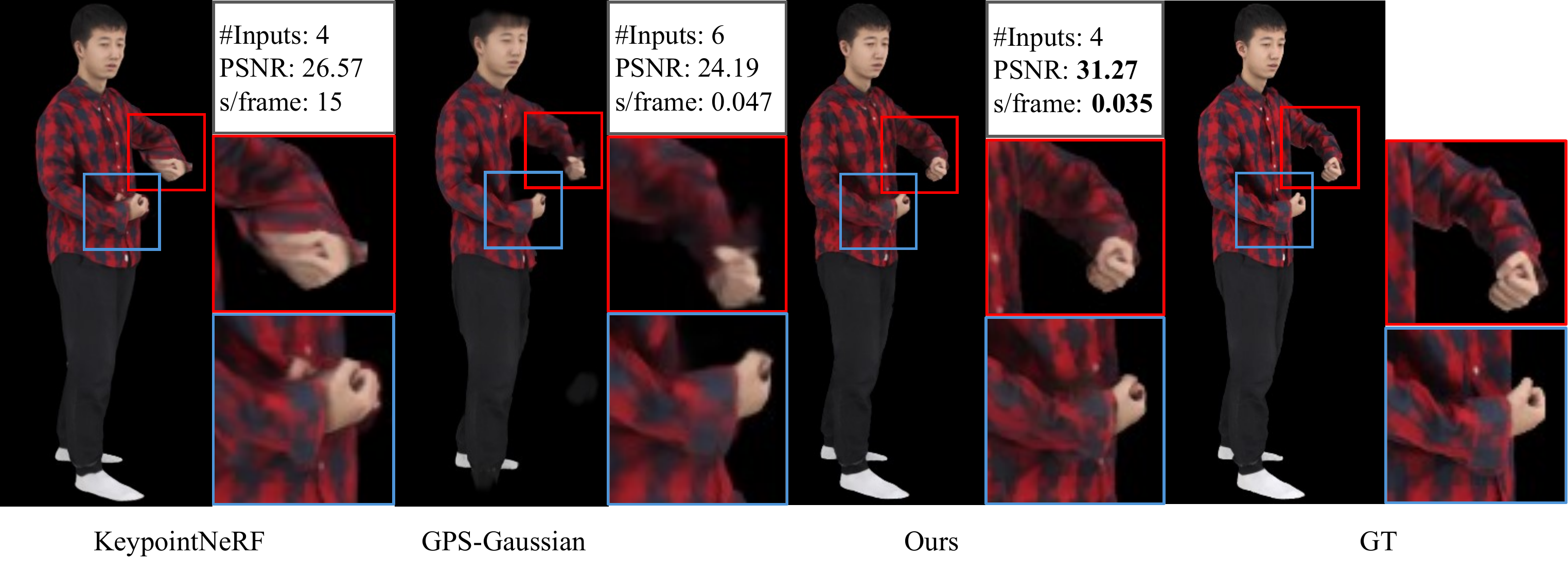}
\vspace{-2em}
\captionof{figure}{Our method enables efficient and high-quality rendering of generalizable human NeRF from sparse input views. Our method outperforms the state-of-the-art quality-prioritized methods, e.g., KeypointNeRF~\cite{keypointnerf}, and has competitive rendering speed with the fastest speed-prioritized methods, e.g., GPS-Gaussian~\cite{gps}. Our method also removes artifacts caused by occlusion in the input views, often observed in state-of-the-art methods. As shown in the figure,
existing methods \knmrA{yield} blurriness and missing body parts
(red boxes) due to using sparse input views, and finger-like artifacts near the human's right hand (blue boxes) due to occlusion. In contrast, our method renders high-quality images without artifacts.
The metric score and rendering speed \knmrA{in each figure} are averaged \knmrA{values} over the test set in our experiments. \vspace{1em}}
\label{fig:teaser}
}]
\begin{abstract}
Generalizable \knmrA{n}eural \knmrA{r}adiance \knmrA{f}ield (NeRF) enables neural-based digital human rendering without per-scene retraining. When combined with human prior knowledge, high-quality human rendering can be achieved even with sparse input views. However, the inference of these methods is still slow, as a large number of neural network queries on each ray are required to ensure the rendering quality. Moreover, occluded regions often suffer from artifacts, especially when the input views are sparse.
To address these issues, we propose a generalizable human NeRF framework that achieves high-quality and real-time rendering with sparse input views by extensively leveraging human prior knowledge. We accelerate the rendering with a two-stage sampling reduction strategy: first constructing boundary meshes around the human geometry to reduce the number of ray samples for sampling guidance regression, and then \knmrA{volume} rendering using fewer guided samples. To improve rendering quality, especially in occluded regions, we propose an occlusion-aware attention mechanism to extract occlusion information from the human priors, followed by an image space refinement network to improve rendering quality. Furthermore,
\wA{for \knmrA{volume} rendering, we \knmrA{adopt} a signed ray distance function (SRDF) formulation,}
which allows us to propose an SRDF loss at every sample position to improve the rendering quality further.
Our experiments demonstrate that our method outperforms the state-of-the-art methods in rendering quality and has a competitive rendering speed compared with speed-prioritized novel view synthesis methods.
\end{abstract}
\begin{table*}[ht]
  \centering
  \caption{A comparison of
  the state-of-the-art methods \knmrA{and our method}
  \knmrA{in terms of} quality-focused \knmrA{and} speed-focused design\knmrA{s}.
  The checkmark (\textcolor{green}{\cmark}) and cross (\xmark) indicate
  \knmrA{whether the corresponding item is applicable to each method or not, respectively.}
  Since GPS-Gaussian~\cite{gps} uses rasterization-based\knmrA{,}
  the ray and sample reduction are not applicable.}
  \setlength{\tabcolsep}{3pt} 
  \label{tab:method_comparison}
  \begin{tabular}{l|cc|cc}
    \toprule
    & \multicolumn{2}{c|}{\textbf{Quality-focused design}} & \multicolumn{2}{c}{\textbf{Speed-focused design}} \\
    \midrule
    \backslashbox{Method}{Design} & Human prior & Occlusion awareness & Ray reduction & Sample reduction \\
    \midrule
    IBRNet~\cite{ibrnet} & \xmark & \xmark & \xmark & \xmark \\
    KeypointNeRF~\cite{keypointnerf} & \color{green}{\cmark} & \xmark & \xmark & \xmark \\
    GM-NeRF~\cite{gmnerf} & \color{green}{\cmark} & \xmark & \xmark & \color{green}{\cmark} \\
    GP-NeRF~\cite{gpnerf} & \color{green}{\cmark} & \xmark & \color{green}{\cmark} & \xmark \\
    ENeRF~\cite{enerf} & \xmark & \xmark & \color{green}{\cmark} & \xmark \\
    GPS-Gaussian~\cite{gps} & \xmark & \xmark & - & - \\
    Ours & \color{green}{\cmark} & \color{green}{\cmark} & \color{green}{\cmark} & \color{green}{\cmark} \\
    \bottomrule
  \end{tabular}
\end{table*}
\section{Introduction}
\label{sec:intro}
Digital humans have a wide range of applications, such as telepresence, virtual/augmented reality, and movie production.
In recent years, methods based on neural implicit representation (e.g., \knmrA{n}eural \knmrA{r}adiance \knmrA{f}ield (NeRF) and its extensions~\cite{nerf,nerfpp,mipnerf}) are utilized for high-quality 3D human rendering.
\knmrA{Using human body} keypoints and statistical human models such as \knmrA{SMPL~\cite{smpl}} and \knmrA{SMPL-X~\cite{smplx}} as prior knowledge,
the rendering quality is further improved, and the number of input views required is also reduced~\cite{neuralbody,humannerf}.
Moreover, to overcome the lengthy per-scene retraining of the original NeRF, generalizable human NeRFs~\cite{gnp, nhp, keypointnerf, gmnerf} utilize pixel-aligned features from input views to enable retraining-free rendering with sparse input views.

However, we find two main issues on rendering speed and quality in the current state-of-the-art generalizable human NeRFs that handle sparse input views.
(1) On rendering speed, the inference time of these methods is still slow, as they require dense neural network queries on each ray to render high-quality images.
Methods that accelerate the rendering of generalizable human NeRFs have been proposed~\cite{gpnerf, enerf, gps}, while they often ignore the human prior knowledge,
\knmrA{deteriorating the} rendering quality on novel subjects,
especially when the input views are sparse.
(2) On rendering quality, occluded regions often suffer from artifacts\knmrA{, especially} when the input views are sparse,
as larger regions are not observable from any input views.
While recent work~\cite{reconfusion,idnerf} has utilized pretrained generative model as prior knowledge to improve rendering quality
\knmrA{for the sparse view setting,}
they either suffer from rendering speed slowdown by using heavyweight models or do not explicitly handle occluded regions.

To address the above issues, we \knmrA{propose EG-HumanNeRF, aiming at accelerating} the rendering of generalizable human NeRFs while improving the rendering quality,
especially in occluded regions under a sparse input view setting.
We achieve our goals by \knmrA{further exploiting} human prior knowledge than previous methods to reduce neural network queries,
provide geometry guidance, and hint for occlusion information.
To accelerate the rendering, we propose a two-stage sampling reduction strategy: (1) In the first stage, we construct boundary meshes around the human geometry and perform the first-stage sampling only within the boundary meshes to extract geometry features. (2) In the second stage, from the geometry features, the sampling range is further narrowed down by regressing sampling guidance based on signed ray distances function (SRDF)~\cite{srdf}.
\wA{SRDF is defined as the \knmrA{signed distance to the nearest surface} along \knmrA{a ray},
which allows us to regress a narrow sampling range for rendering.}
The actual neural rendering in the second stage only needs a few samples without rendering quality degradation.
To improve the rendering quality in a sparse input setting, especially in occluded regions, we propose an occlusion-aware attention mechanism to hint the network for occlusion region locations. An image space refinement network utilizes the regressed occlusion information to improve rendering quality in occluded regions without introducing heavyweight models. The image space refinement network also performs upsampling, which reduces the number of rays to sample and accelerates the rendering process further.
Furthermore, thanks to the
\wA{SRDF formulation}
 we propose an SRDF loss to supervise the predicted geometry at every sample position when the ground\knmrA{-}truth geometry is available, further improving the rendering quality.

The experiments in Sec.~\ref{sec:exp} demonstrate that our method outperforms the state-of-the-art methods in terms of rendering quality and has competitive rendering speed with the fastest NeRF-based and even 3D Gaussian-based methods, as illustrated in Fig.~\ref{fig:teaser}. A comparison of our method with the state-of-the-art methods on designs for rendering quality and speed is summarized in Table~\ref{tab:method_comparison}.
In summary, our main contributions are as follows:
\begin{itemize}
    \item We propose a generalizable human NeRF framework that utilizes human prior knowledge for both high-quality and efficient rendering using as sparse as three to four input views.
    \item Utilizing the SMPL-X mesh as human prior knowledge, we reduce the number of ray samples requiring neural network inference in a two-stage manner. Specifically, we reduce the ray samples for both sampling guidance regression and \knmrA{volume} rendering, which significantly accelerates the rendering process.
    \item We predict occluded regions in the target view with human prior knowledge and use image space refinement to improve rendering quality in occluded regions.
    \item By using an
    \wA{SRDF formulation for rendering,}
    we propose an SRDF loss at every sample position to further improve the rendering quality.
\end{itemize}
\section{Related Work}
\label{sec:related}
\subsection{Neural Implicit Representation for Human}
Traditional 3D construction methods for digital human require dense camera arrays~\cite{densecam_mesh_08, densecam_mesh_09, densecam_volume} or depth sensors~\cite{depth_multi, depth_single_20, depth_single_22} for mesh or volume-based geometry reconstruction. These methods, however, either give unsatisfactory rendering quality because of the limited geometry reconstruction accuracy or require costly hardware setups, hindering their practical use.
Neural implicit representation is a promising alternative for representing 3D scenes~\cite{nerf,siren,neus,volrecon}, including full-body human~\cite{neuralbody,animnerf,humannerf,anerf,doublefield} because it enables high-quality rendering without explicit geometry reconstruction.
One of the most widely adopted neural implicit representations is \knmrA{n}eural \knmrA{r}adiance \knmrA{f}ield (NeRF). NeRF represents the 3D scene as a volumetric radiance field where view-dependent radiances at 3D positions are encoded by \knmrA{m}ulti-layer \knmrA{p}erceptrons (MLPs). While yielding high-quality rendering results, NeRF requires lengthy neural network training for individual scenes.
Follow-up work~\cite{ibrnet, mvsnerf, pixelnerf} extends NeRF so that it generalizes on unseen scenes given sparse input views. However, these regression-based methods for general scenes often perform worse than the optimization-based methods unless conducting a post hoc fine-tuning for the specific scene~\cite{neus, volrecon}.
For humans, prior knowledge of human pose and body shape, such as keypoints and statistical models~\cite{smpl, smplx}, can be used to guide the view synthesis~\cite{gnp, nhp, keypointnerf, gmnerf}, improving rendering quality on novel subjects. Nevertheless, the inference time of these methods is still slow, as they require a large number of neural network queries on each ray to render high-quality images.
Our method accelerates the rendering of generalizable human NeRFs while utilizing prior knowledge, enabling both efficient and high-quality rendering with sparse inputs.

\wA{Existing work~\cite{dietnerf,infonerf,manifoldnerf} studied on improving the rendering quality for NeRF when using sparse inputs.
A specific issue when using sparse views for novel view synthesis is that regions non-observable from inputs may contain artifacts.}
Pretrained generative models are often utilized to hallucinate the occluded regions. ReconFusion~\cite{reconfusion} trains a diffusion model for generalizable novel view synthesis, which is then used to guide the training of an optimization-based NeRF on occluded regions. The heavyweight diffusion model, however, is unsuitable for retraining-free real-time rendering. While ID-NeRF~\cite{idnerf} distills a pretrained diffusion model into a generalizable NeRF to improve rendering quality with sparse inputs, the guidance from the diffusion model applies only to the input view features, making its ability to hallucinate occluded regions questionable.
To avoid slowing down the rendering process, our method, on the other hand, first extracts occlusion information from human prior knowledge, making the subsequent image space refinement for occluded regions efficient and lightweight.

\begin{figure*}[ht]
  \centering
  \includegraphics[width=1\linewidth]{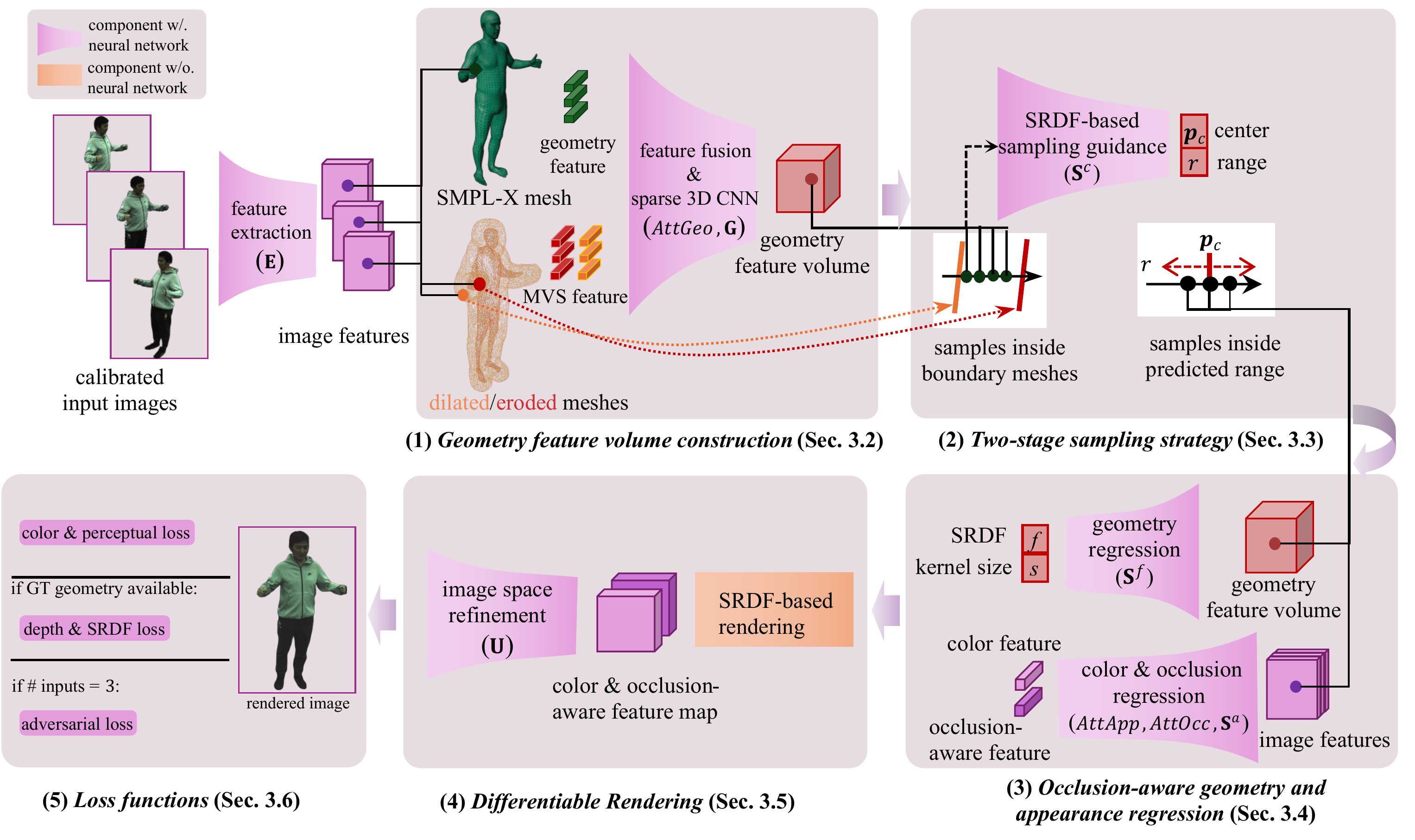}
  \caption{\wA{Overview of our method.
  Given calibrated sparse multi-view images\knmrA{,} we \knmrA{encode their feature maps and} (1) construct a geometry feature volume from the SMPL-X mesh and the input views to provide human prior knowledge.
  Utilizing the boundary meshes derived from the SMPL\knmrA{-X} mesh, we (2) use a two-stage sampling strategy to reduce the number of samples required and accelerate the rendering while maintaining rendering quality. \knmrA{At} each sample position, we (3) regress geometry-related values and appearance features used for rendering, along with an optional occlusion-aware feature to improve rendering quality in occluded regions.
  We aggregate features on sample positions using an SRDF-based formulation.
  Finally, we (4) conduct image space refinement from feature maps to synthesize the image in \knmrA{the} target view. Our method (5) is trained end-to-end, with optional SRDF and adversarial loss\knmrA{es} to improve the rendering quality further.}}
  \label{fig:method}
\end{figure*}

\subsection{Neural Volume Rendering Acceleration}
Variant methods have been proposed to accelerate the rendering of NeRF. A few strategies for accelerating both NeRF training and inference include representing the scene with a sparse feature field and shallow MLPs instead of deep MLPs to save computation on each ray sample~\cite{voxel,direct,octree,ngp,tenso,eg3d}, using image space upsampling to reduce the number of rays to calculate~\cite{stylenerf,mobiler2l,duplex}, and using efficient sampling strategies to reduce the number of samples~\cite{donerf,efficient,adanerf}. Some methods focus on inference (rendering) acceleration by baking the outputs of the neural network into a sparse feature volume~\cite{snerg,merf} or polygons~\cite{mobilenerf}, or extracting boundary meshes to restrict the sampling space as in Adaptive Shells~\cite{adashell}.
\wA{Our method also utilizes boundary meshes for efficient sampling, but the fine-grained boundary meshes used in Adaptive Shells~\cite{adashell} are not available for our retraining-free setting; Adaptive Shells first optimizes radiance fields for specific scenes, then extracts boundary meshes.
Instead, we build boundary mesh from the coarse SMPL-X mesh, and use a second-stage sampling range regression to compensate for the coarse boundary mesh and further reduce the number of samples.}

The following works focus on accelerating the rendering of generalizable human NeRF. GP-NeRF~\cite{gpnerf} and ENeRF~\cite{enerf} achieve efficient rendering by reducing the number of sample positions in volume rendering. GP-NeRF constructs a feature volume from SMPL mesh and utilizes the volume to eliminate samples outside the volume.
ENeRF~\cite{enerf} constructs a cost volume to estimate depth that guides the sampling. However, ENeRF does not employ prior knowledge of the human body,
\knmrA{deteriorating the} rendering quality on novel subjects,
especially when the input views are sparse.
GPS-Gaussian~\cite{gps} regresses parameters of 3D Gaussians from the input views to enable fast rendering without neural network inference. However, GPS-Gaussian requires adjacent views to contain fully overlapping subjects for regressing the positions of 3D Gaussians,
making it inapplicable to \knmrA{as sparse as three or four} input views.
Moreover, GPS-Gaussian shares the same limitation as ENeRF in that it does not utilize prior knowledge of the human body, leading to degradation in rendering quality even with denser input views (e.g., six views in our experiments).
\knmrA{A concurrent work conceptually similar to our approach, GHG~\cite{ghg},} utilizes human prior for 3D Gaussian by learning the parameters of the 3D Gaussians in the UV space defined by the SMPL-X model.
\knmrA{While GHG adopts the fast rendering paradigm of volume splatting, we propose an alternative approach for volume rendering.}
\section{Method}
\label{sec:method}
\subsection{Overview}
\knmrA{Fig.~\ref{fig:method} illustrates our} generalizable human NeRF framework\knmrA{.}
From calibrated sparse multi-view images, our method utilizes the SMPL-X parametric model as both the prior knowledge and the sampling guidance to learn a neural implicit representation that can generalize to novel subjects and render in \knmrA{real time}.
To render a novel view, we first construct a geometry feature volume from the SMPL-X mesh and the input views (Sec.~\ref{sec:geom}). The feature volume is sparse and contains human geometry prior only around the human surface, enabling an efficient 3D sparse convolution for feature aggregation.
We use a two-stage sampling strategy so that it only requires a small number of samples to render high-quality images, accelerating the rendering process (Sec.~\ref{sec:sampling}). Specifically, from the geometry feature volume, we first take a small number of ray samples within the boundary meshes derived from the SMPL mesh. With the sampled features, we regress signed ray distances as sampling guidance, further reducing the number of second-stage samples required for \knmrA{volume} rendering.
\knmrA{At} each sample position taken in the second stage, we sample the geometry volume, and input view features encoded by a convolutional neural network (CNN) to regress geometry-related values and appearance features used for rendering (Sec.~\ref{sec:occ_and_reg}). When the unobserved regions in the target view are large (i.e., when three input views are used), we regress occlusion-aware features from the SMPL-X mesh and the input views, which are used in later steps to improve rendering quality in occluded regions.
Finally, we aggregate features on sample positions using an
\wA{SRDF formulation}
followed by a 2D CNN to refine and render the final image (Sec.~\ref{sec:render}). When the occlusion-aware features are available, we predict the occlusion mask to aid the refinement of occluded regions. The 2D CNN also serves as an upsampler to reduce the number of rays to calculate before convolutions, thus further accelerating the rendering process.
Our method is trained end-to-end with a color and a perceptual loss. When the ground\knmrA{-}truth geometry is available, we also supervise the predicted geometry with an SRDF loss to further improve the rendering quality (Sec.~\ref{sec:loss}). When the occlusion-aware refinement is enabled, we use an additional adversarial loss along with a discriminator to fine-tune a trained model, improving the rendering quality in occluded regions.

\subsection{Geometry Feature Volume Construction} \label{sec:geom}
To provide prior knowledge on human geometry and guidance for efficient sampling,
we construct a geometry feature volume by aggregating multi-view image features and human geometry prior from the SMPL-X mesh\knmrA{.}
\wA{Our method starts from \(N_I\) calibrated multi-view images \(\{I_i \in \mathbb{R}^{H \times W \times 3} \}_{i=1}^{N_I}\) \knmrA{(where $H$ and $W$ are the image height and width)} of a human subject and a fitted SMPL-X mesh with vertices \(\{\mathbf{v}_j \in \mathbb{R}^{3} \}_{j=1}^{N_v}\).}
From the SMPL-X mesh, we construct a pair of dilated and eroded meshes consisting of mesh vertices \(\{\mathbf{v}^d_j \in \mathbb{R}^{3}\}_{j=1}^{N_v}\) and \(\{\mathbf{v}^e_j \in \mathbb{R}^{3}\}_{j=1}^{N_v}\), respectively,
by expanding and shrinking the \knmrA{mesh} along the normal direction\knmrA{s}
\wA{to a fixed distance.}
The pair of boundary meshes are used for geometry feature volume construction and sampling guidance.
\wA{Specifically, for the case of a ray intersecting the both dilated and eroded meshes, we sample uniformly between the first intersections with the dilated and eroded meshes.
For the case of a ray intersecting only the dilated mesh, we sample uniformly between the first and last intersections with the dilated mesh.}
We first extract \(N_I\) feature maps \(\{I^h_i \in \mathbb{R}^{H^h \times W^h \times C^{h}}\}_{i=1}^{N_I}\) \knmrA{(where $H^h$, $W^h$, and $C^h$ are the height, width, and number of channels of the feature maps, respectively)} from the input images using a pretrained CNN \(\mathbf{E}\) based on EfficientNet~\cite{efficientnetv2}. Given a 3D position \(\mathbf{p} \in \mathbb{R}^3\) and the feature map \(I^h_i\), we can project \(\mathbf{p}\) to the image plane of an input view \knmrA{via} \wA{coordinate} transformation from the world coordinate to the input camera coordinate
\(\Pi_i(\cdot)\)\knmrA{,} followed by a bi-linear interpolation \(\Phi_{\mathit{bi}}(\cdot)\). Namely, a pixel-aligned feature \(\mathbf{h}_i(\mathbf{p})\) is obtained by
\begin{equation}
    \mathbf{h}_i(\mathbf{p}) = \Phi_{\mathit{bi}}(I^h_i, \Pi_i(\mathbf{p})).
\end{equation}
Substituting \(\mathbf{p}\) with the mesh vertex positions \(\mathbf{v}_j\), \(\mathbf{v}^d_j\), and \(\mathbf{v}^e_j\), we obtain the features \(\mathbf{h}_i(\mathbf{v}_j)\), \(\mathbf{h}_i(\mathbf{v}^d_j)\), and \(\mathbf{h}_i(\mathbf{v}^e_j)\) at vertices of SMPL-X, dilated, and eroded meshes, respectively.
To provide sufficient information for the neural network to learn the human geometry, we aggregate the multi-view image feature
\( \{\mathbf{h}_i(\mathbf{v}_j)\}_{i=1}^{N_I}\)
along with the human geometry prior, namely, the normal direction \(\mathbf{n}_j\) of each SMPL-X vertex using the attention mechanism in GM-NeRF~\cite{gmnerf} (\knmrA{i.e.,} Eq.~(3) \knmrA{in the paper}) as \(\mathbf{a}(\mathbf{v}_j) = \operatorname{AttGeo}(\{\mathbf{h}_i(\mathbf{v}_j), \mathbf{d}_i(\mathbf{v}_j)\}_{i=1}^{N_I}, \mathbf{n}_j)\), where \(\mathbf{d}_i(\mathbf{v}_j)\) is the direction from the \(i\)\knmrA{-th} input camera position to the vertex \(\mathbf{v}_j\).
Notice that feature \(\mathbf{a}(\mathbf{v}_j)\) only provides the geometry information on each SMPL-X vertex, which can be inaccurate. To provide denser information around the human body, we further provide multi-view stereo information around the surface of the human body.
Specifically, we aggregate the multi-view image feature \(\mathbf{h}_i(\mathbf{v}^d_j)\) and \(\mathbf{h}_i(\mathbf{v}^e_j)\) by calculating their mean and variance of input views and concatenating them to produce \(\mathbf{m}(\mathbf{v}^d_j)\) and \(\mathbf{m}(\mathbf{v}^e_j)\).
\wA{The features} \(\{\mathbf{a}(\mathbf{v}_j), \mathbf{m}(\mathbf{v}^d_j), \mathbf{m}(\mathbf{v}^e_j)\}_{j=1}^{N_v}\) are then fed into a sparse 3D convolutional network \(\mathbf{G}\) to construct the geometry feature volume \(\wA{\mathcal{G}} \in \mathbb{R}^{D_\mathcal{G} \times H_\mathcal{G} \times W_\mathcal{G}  \times C_\mathcal{G}}\), where \(D_\mathcal{G}\), \(H_\mathcal{G}\), \(W_\mathcal{G}\), and \(C_\mathcal{G}\) are the depth, height, width, and \knmrA{the number of} channel\knmrA{s} of the feature volume, respectively.

\subsection{Two-stage Sampling Strategy} \label{sec:sampling}
We propose a two-stage sampling strategy to reduce the total number of samples that require neural network inference, accelerating the rendering process while maintaining rendering quality.
In the first stage, we sample the geometry feature volume on each ray to regress the sampling guidance for the second stage. Instead of sampling inside the scene bounding box, we utilize the dilated and eroded boundary meshes constructed to sample within, skipping the empty space.
Formally, given a ray \(\mathbf{r}\) and the geometry feature volume \(\mathcal{G}\), we first uniformly sample a set of 3D positions \(\{\mathbf{p}_k\}_{k=1}^K\) inside the boundary meshes and extract the corresponding geometry feature \(\mathbf{g}(\mathbf{p}_k) = \Phi_{\mathit{tri}}(\knmrA{\mathcal{G}}, \mathbf{p}_k)\), where \( \Phi_{\mathit{tri}}(\cdot)\) is the tri-linear interpolation operator.
In the second stage, to further reduce the number of samples required for rendering,
\wA{
we specify the sampling range by regressing a sampling center \knmrA{ \(\mathbf{p}_c\)} and sampling radius \knmrA{$r$},}
for each ray using
\wA{an SRDF formulation.}
From the geometry features \(\mathbf{g}_k\) and the ray direction \(\mathbf{d}\), we regress the signed ray distances \(\{f_k\}_{k=1}^K\) along with a confidence score \(\{c_k\}_{k=1}^K\) using a shallow MLP \(\mathbf{S}^c\)\knmrA{:}
\begin{equation} \label{eq:mlp_c}
    \begin{split}
    (f_k, c_k) = S^c(\mathbf{g}_k, \mathbf{d}).
    \end{split}
\end{equation}
The sampling center \(\mathbf{p}_c\) is then calculated as
\knmrA{a weighted average}
of the \knmrA{3D positions} calculated from the SRDFs\knmrA{:}
\begin{equation}
    \mathbf{p}_c = \knmrA{\frac{1}{\sum_{k=1}^K c_k}} \sum_{k=1}^K c_i (\mathbf{p}_o + f_k \mathbf{d}),
\end{equation}
where \(\mathbf{p}_o\) is the camera position \knmrA{of the target view}. The sampling radius \(r\) is calculated as the standard deviation of
\knmrA{weighted distances between the 3D positions and $\mathbf{p}_c$:}
\begin{equation}
    r = \left({\knmrA{\frac{1}{\sum_{k=1}^K c_k}} \sum_{k=1}^K c_k (\mathbf{p}_o + f_k \mathbf{d} - \mathbf{p}_c)^2} \right)^{\frac{1}{2}}.
\end{equation}
The advantage of predicting SRDFs instead of the
\wA{ray parameter}
directly is that we can use an SRDF loss (Sec.~\ref{sec:loss}) to supervise the SRDFs at every sample position, which helps further improve the rendering quality when the ground\knmrA{-}truth geometry is available.
We uniformly sample \(L\) 3D positions \(\{\mathbf{p}_{l} = \mathbf{p}_{c} + t_{l} \mathbf{d}\}_{l=1}^{L}\), where \(t_l \in [-r, r]\).
Namely, we sample 3D positions
\knmrA{within radius $r$ from center $\mathbf{p}_c$}
to render the novel view.

With the strong geometry guidance provided by the boundary meshes and the geometry feature volume, it only requires a small number of \(L\) samples to render high-quality images.

\subsection{Occlusion-aware Geometry and Appearance Regression} \label{sec:occ_and_reg}
Given the guided sample positions, we regress geometry and appearance values or features at sample positions for rendering, which we describe in Sec.~\ref{sec:regression}.

Different from previous work, we optionally use an attention mechanism to regress an occlusion-aware feature (Sec.~\ref{sec:occlusion}) from the SMPL-X geometry and the input views when the unobserved regions in the target view are large.
The occlusion-aware features contain information on whether the input views are reliable for the sample position, which is used to regress an occlusion mask, aiding the image space refinement on occluded regions (Sec.~\ref{sec:image_space}).
\begin{figure*}[ht]
  \centering
  \includegraphics[width=1\linewidth]{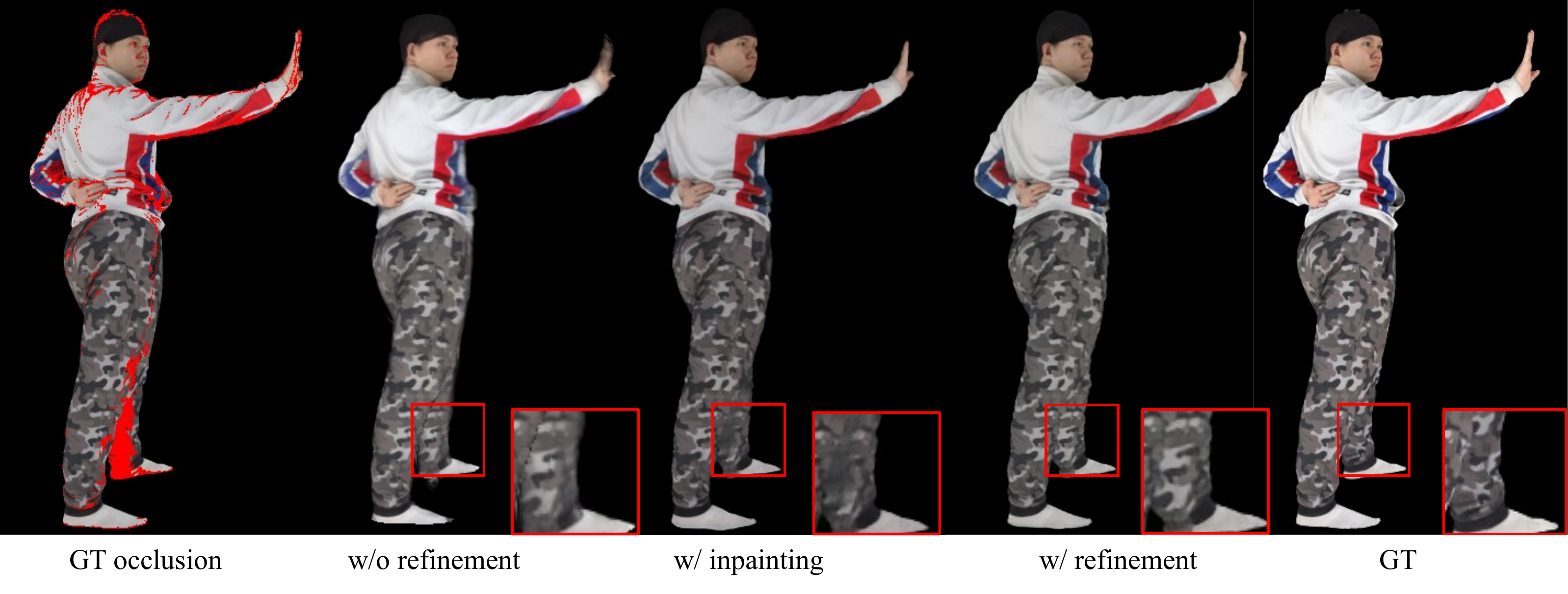}
  \caption{Comparisons between inpainting and image space refinement. ``w/o refinement'' denotes the rendering results without using image space \wA{refinement}. ``w/ inpainting'' denotes rendering results with a post hoc inpainting method~\cite{lama}. ``w/ refinement'' denotes our rendering results with image space refinement using occlusion-aware features.
  When the input views already provide sufficient information to hallucinate the occluded regions, the inpainting method destroys the original information and is not able to recover complex textures without using a heavyweight model. In contrast, the image space refinement method focus on the occluded regions, preserving the original information and avoiding the need for heavyweight generative models.}
  \label{fig:inp_or_ref}
\end{figure*}

\subsubsection{Geometry and Appearance Regression} \label{sec:regression}
Given sample positions \(\{\mathbf{p}_l\}_{l=1}^L\), we sample the geometry feature volume \(\mathcal{G}\) and the image feature maps \(\{I^h_i\}_{i=1}^{N_I}\) at the sample positions to obtain geometry features \(\mathbf{g}(\mathbf{p}_l) = \Phi_{\mathit{tri}}(\mathcal{G}, \mathbf{p}_l)\) and image features \(\{\mathbf{h}_i(\mathbf{p}_l)\}_{i=1}^{N_I}\), with \(\mathbf{h}_i(\mathbf{p}_l)=\Phi_{\mathit{bi}}(I^h_i, \Pi_i(\mathbf{p}_l))\). From the sampled features, geometry-related values, and appearance features, along with colors and occlusion-aware features, are regressed.

Similar to Eq.~\eqref{eq:mlp_c}, we use another shallow MLP \(\mathbf{S}^f\) to regress the geometry-related values \((f_l, s_l) = S^f(\mathbf{g}(\mathbf{p}_l), \mathbf{d})\). The outputs include an SRDF value \(f_l\) and an additional kernel size \(s_l\), which will be used for the rendering process described in Sec.~\ref{sec:render}.

To blend the image features from the input views while considering the geometry, we utilize the attention mechanism in GM-NeRF~\cite{gmnerf} (\knmrA{i.e.,} Eq.~(5) \knmrA{in the paper}) to obtain the appearance features \(\mathbf{a}(\mathbf{p}_l) = \operatorname{AttApp}(\{\mathbf{h}_i(\mathbf{p}_l), \mathbf{r}_i(\mathbf{p}_l), \mathbf{d}_i(\mathbf{p}_l)\}_{i=1}^{N_I}, \mathbf{g}(\mathbf{p}_l), \mathbf{d}(\mathbf{p}_l))\), where \(\mathbf{r}_i\), \(\mathbf{d}_i\), and \(\mathbf{d}\) are the projected input views raw pixels from \(\mathbf{p}_l\), the input camera ray direction, and the target camera ray direction at \(\mathbf{p}_l\), respectively. We subsequently use another shallow MLP \(\mathbf{S}^a\) to regress the RGB color \(\mathbf{c}_l\) from the appearance feature \(\mathbf{a}(\mathbf{p}_l)\).

\subsubsection{Occlusion-aware Attention} \label{sec:occlusion}
In order to improve the rendering quality in occluded regions with a limited model capacity, it is crucial to provide guidance on the exact occlusion regions so that the network can know for which pixels the input view information may not be reliable and should be hallucinated instead of copied from input views. To this end, we propose an occlusion-aware attention mechanism that calculates features indicating whether occlusions exist in input views given a sample position \(\mathbf{p}\). We first render a world-space position map \(X'\) of the SMPL-X mesh from the target view and position maps \(\{X_i\}_{i=1}^{N_I}\) of the SMPL-X mesh from the input views. We then calculate the distance \(d_i\) between \(\mathbf{p}\)'s observed positions in the target view and in the \(i\)\knmrA{-th} input view as follows:
\begin{equation}
  \begin{split}
  \mathbf{x}'(\mathbf{p}) = \Phi_{\mathit{bi}}(X', \Pi'(\mathbf{p})), \\
  {\mathbf{x}}_i(\mathbf{p}) = \Phi_{\mathit{bi}}(X_i, \Pi_i(\mathbf{p})), \\
  d_i(\mathbf{p}) = \|\mathbf{x'}(\mathbf{p}) - \mathbf{x}_i(\mathbf{p})\|_2.
  \end{split}
\end{equation}
The distance \(d_i\) \knmrA{should be (close to)} zero when the target and the \(i\)\knmrA{-th} input view observe the same position and large when they observe different positions, indicating occlusions. Our occlusion-aware attention that calculates the occlusion-aware feature \(\mathbf{o}(\mathbf{p})
=\operatorname{AttOcc}(\{\mathbf{g}(\mathbf{p}), \mathbf{d}(\mathbf{p}), \{d_i(\mathbf{p})\}_{i=1}^{N_I}, \mathbf{x}'(\mathbf{p}), \{\mathbf{x}_i(\mathbf{p})\}_{i=1}^{N_I}\})
\) is then formulated as follows:
\begin{equation} \label{eq:occlusion}
  \begin{split}
  \mathbf{Q}_o(\mathbf{p}) \! &= \! F_o^Q(\mathbf{g}(\mathbf{p}) \! \oplus \! \mathbf{d}(\mathbf{p}) \! \oplus \! {\mathbf{x}'} \!(\mathbf{p})), \\
  \mathbf{K}_o(\mathbf{p}) \! &= \! F_o^K \!\! \left( \! \left\{\mathbf{h}_i(\mathbf{p}) \! \oplus \! \mathbf{d}_i(\mathbf{p}) \! \oplus \! {\mathbf{x}}_i(\mathbf{p}) \! \oplus \! d_i(\mathbf{p}) \right\}_{i=1}^{N_I} \! \right), \\
  \mathbf{V}_o(\mathbf{p}) \! &= \! F_o^V \!\! \left( \! \left\{\mathbf{g}(\mathbf{p}) \! \oplus \! d_i(\mathbf{p}) \right\}_{i=1}^{N_I} \! \right), \\
  \mathbf{o}(\mathbf{p}) \!  &= \! F_o \! \left(\operatorname{Att}\left(\mathbf{Q}_o(\mathbf{p}), \mathbf{K}_o(\mathbf{p}), \mathbf{V}_o(\mathbf{p})\right)\right),
  \end{split}
\end{equation}
where \(\oplus\) denotes the concatenation operation, \(F_o^Q\), \(F_o^K\), \(F_o^V\), and \(F_o\) are linear layers that produce the query, key, value, and output features, respectively. The intuition behind Eq.~\eqref{eq:occlusion} is that, while whether \knmrA{$\mathbf{p}$} is occluded or not can be estimated simply from \(\min\{d_i\}_{i=1}^{N_I}\), the SMPL-X mesh differs from the ground\knmrA{-}truth human geometry, causing inaccuracies in the occlusion estimation. The attention mechanism,
however, learns to decide whether the SMPL-X geometry is reliable or not for the occlusion estimation from the geometry feature \(\mathbf{g}\), image feature \(\mathbf{h}_i\), and the input and target view direction \(\mathbf{d}_i\) and \(\mathbf{d}\). This attention mechanism is proven to be effective for robustly blending multi-view features in GM-NeRF~\cite{gmnerf} when the SMPL-X geometry is inaccurate; thus, we adapt this mechanism to handle occlusion-aware features. We optionally calculate the occlusion-aware features, which will be used in the image space refinement (Sec.~\ref{sec:image_space}) when the unobserved regions in the target view are large, i.e., when three input views are used in our experiment.

\subsection{Differentiable Rendering} \label{sec:render}
To render the novel-view image, we first render a low-resolution feature map using the SRDF-based \knmrA{volume} rendering method in VolRecon~\cite{volrecon}. The feature map is fed into a 2D CNN to predict the final full-resolution image\knmrA{.} 
The purpose of the 2D CNN is two-fold: (1) to reduce calculations that depend on \knmrA{the} sample number by reducing the number of rays to sample, and (2) to refine regions fully occluded in input views.
\begin{figure*}[ht]
  \centering
  \includegraphics[width=1\linewidth]{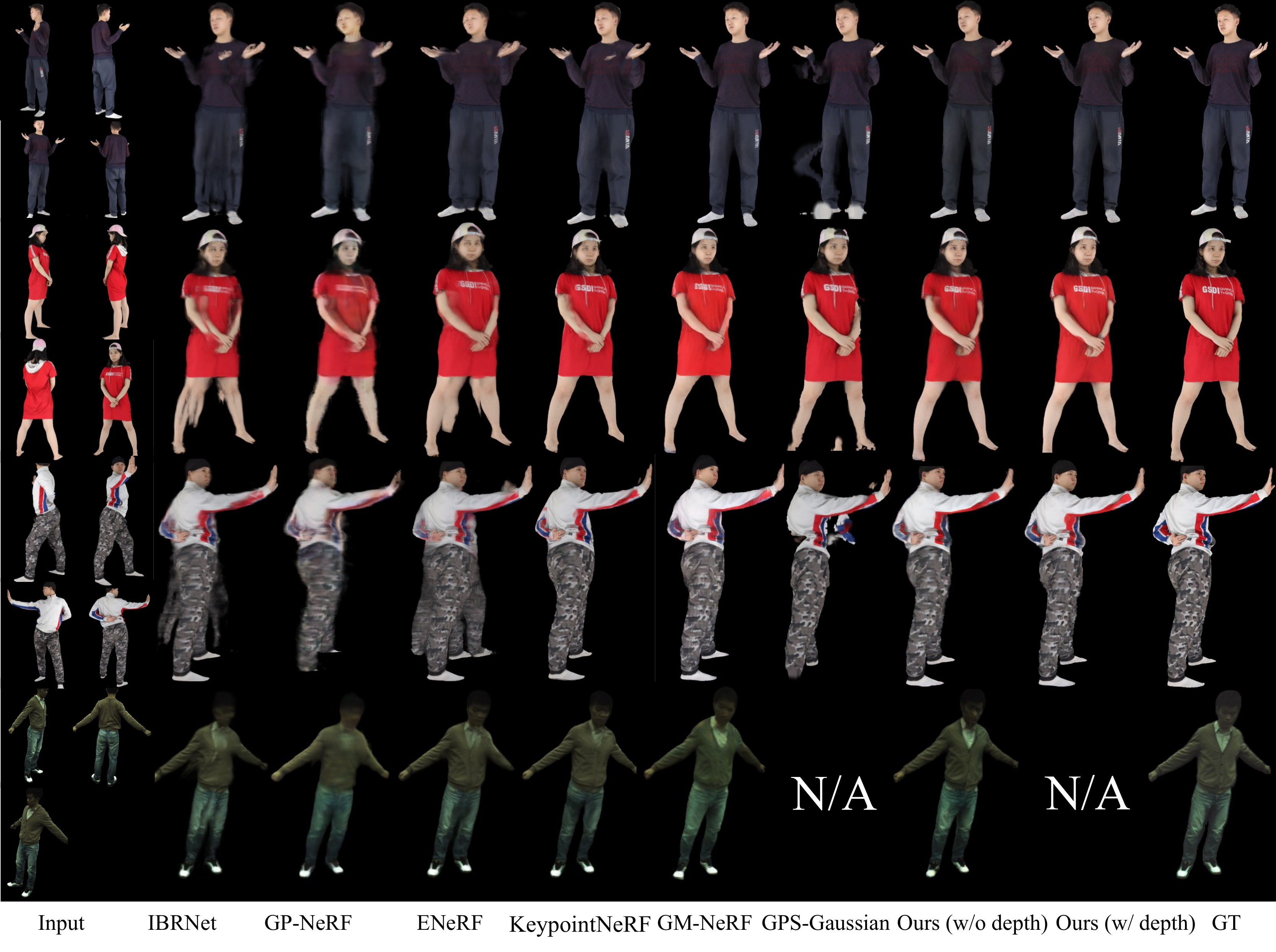}
  \caption{Qualitative evaluation results on the THuman2.0 dataset \knmrA{(first three rows)} and the ZJU-MoCap dataset \knmrA{(last row)} compared with the state-of-the-art methods.
  We use four input views for the THuman2.0 dataset and three input views for the ZJU-MoCap dataset.
  \knmrA{For GPS-Gaussian~\cite{gps}, we use six input views and ground-truth depth instead of four, as six is the minimum number supported by this method.}
  }
  \label{fig:eval_qual}
\end{figure*}

\subsubsection{Volumetric Rendering}
Given a feature vector \(\mathbf{f}_l\) (or a scalar value) at the sample position \(\mathbf{p}_l\), we aggregate the features as follows:
\begin{equation} \label{eq:render}
    \hat{\mathbf{f}}=\sum_{l=1}^L T_l \, {\alpha}_l \, \mathbf{f}_l ,
\end{equation}
where \(T_l=\prod_{n=1}^{l-1}\left(1-\alpha_n\right)\) is the transmittance and \(\alpha_l\) is the opacity defined as \(\alpha_l=1-\exp \left(-\int_{t_l}^{t_{l+1}} \sigma(t) d t\right)\), with \(t_l\) being the distance from the camera to the \(l\)'s sample position. To obtain opaque density \(\sigma(t)\) from SRDF \(f\), we adopt the formulation from NeuS~\cite{neus} and replace the signed distance function with SRDF as done in VolRecon~\cite{volrecon} as follows:
\begin{equation}
  \begin{split}
  \sigma(t)=\max \left(-\frac{\frac{d \Phi_s}{d t}(f)}{\Phi_s(f)}, 0\right), \\
  \Phi_s(f)=(1+\exp (-f / s))^{-1},
  \end{split}
\end{equation}
where \(s\) is the learnable kernel size controlling the sharpness of the density function. Please refer to VolRecon~\cite{volrecon} and NeuS~\cite{neus} for more theoretical details.
We trace the rays at pixels for an image with half of the target resolution and render the low-resolution feature map \(\hat{I}_a \in \mathbb{R}^{\frac{H'}{2} \times \frac{W'}{2} \times C_a}\) and \(\hat{I}_o \in \mathbb{R}^{\frac{H'}{2} \times \frac{W'}{2} \times C_o}\) for the appearance and occlusion-aware features by substituting \(\mathbf{f}_l\) with \(\mathbf{a}_l\) and \(\mathbf{o}_l\) in Eq.~\eqref{eq:render}, respectively.
\wA{\(H'\) and \(W'\) are the height and width of the target image, respectively.}
We also calculate a low-resolution RGB and depth image \(\hat{I}_\mathit{low} \in \mathbb{R}^{\frac{H'}{2} \times \frac{W'}{2} \times 3}\) and \(\hat{I}_d \in \mathbb{R}^{\frac{H'}{2} \times \frac{W'}{2}}\) by substituting \(\mathbf{f}_l\) with \(\mathbf{c}_l\) and \(t_l\).

\subsubsection{Occlusion-aware Image Space Rendering} \label{sec:image_space}
We use a 2D CNN \(\mathbf{U}\) to predict the final full-resolution image \(\hat{I} \in \mathbb{R}^{H \times W \times 3}\) from the low-resolution feature map\knmrA{s} \(\hat{I}_a\), \(\hat{I}_o\), and the RGB image \(\hat{I}_{low}\). Specifically, we first use a shallow 2D CNN \(\mathbf{O}\) to predict the occlusion mask \(\hat{O} \in \mathbb{R}^{H \times W}\) from the low-resolution occlusion-aware feature map \(\hat{I}_o\). The full-resolution image is calculated as follows:
\begin{equation}
  \begin{split}
    \hat{I} = \mathbf{U}(\hat{I}_a \oplus \hat{I}_o \oplus\operatorname{Downsample}(\hat{O}, \frac{1}{2})) \\
    + \operatorname{Upsample}(\hat{I}_{low}, 2),
  \end{split}
\end{equation}
where \(\operatorname{Downsample}(\cdot, \frac{1}{2})\) and \(\operatorname{Upsample}(\cdot, 2)\) are the bi-linear downsampling and upsampling operators, respectively.
The CNN \(\mathbf{U}\) not only upsamples and decodes the feature map but also improves the rendering quality, especially on occluded regions using the occlusion mask \(\hat{O}\) as guidance.
\begin{table*}[ht]
  \centering
  \caption{Quantitative comparison with state-of-the-art methods on ZJU-MoCap and THuman2.0 datasets.
  The inference time is measured on a single NVIDIA A6000 GPU, except for IBRNet~\cite{ibrnet}.
  \(*\) We \knmrA{measured} the inference time \knmrA{of} IBRNet~\cite{ibrnet} on \knmrA{an old GPU,} NVIDIA Quadro RTX 8000 GPU, because of the \knmrA{old} runtime \knmrA{setting} of IBRNet's official implementation.
  }
  \label{tab:quality}
  \begin{tabular}{l|ccc|ccc|c}
    \toprule
    & \multicolumn{3}{c|}{ZJU-MoCap} & \multicolumn{3}{c}{THuman2.0} \\
    \midrule
    Method & PSNR \(\uparrow\) & SSIM \(\uparrow\) & LPIPS \(\downarrow\) & PSNR \(\uparrow\) & SSIM \(\uparrow\) & LPIPS \(\downarrow\) & \(\sfrac{\text{ms}}{\text{frame}}\) \\
    \midrule
    IBRNet~\cite{ibrnet} & 25.82 &  0.846 & 0.178 & 27.77 & 0.928 & 0.098 & \(7374.83^*\) \\
    KeypointNeRF~\cite{keypointnerf} & 25.54 & 0.874 & \underline{0.104} & 26.57 & 0.939 & 0.055 & 15097.23\\
    GM-NeRF~\cite{gmnerf} & \underline{26.38} & \underline{0.881} & 0.105 & 28.12 & 0.944 & 0.068 & 253.66 \\
    GP-NeRF~\cite{gpnerf} & 24.94 & 0.863 & 0.137 & 26.78 & 0.918 & 0.111 & 158.63 \\
    ENeRF~\cite{enerf} & 24.54 & 0.857 & 0.134 & 25.66 & 0.898 & 0.100 & \underline{40.33} \\
    GPS-Gaussian~\cite{gps} & N/A & N/A & N/A & 24.19 & 0.921 & 0.077 & 46.97 \\
    Ours (w/o depth) & \textbf{26.85} & \textbf{0.891} & \textbf{0.086} & \underline{29.44} & \underline{0.951} & \underline{0.060} & \textbf{34.95} \\
      Ours (w/ depth) & N/A & N/A & N/A & \textbf{31.73} & \textbf{0.973} & \textbf{0.034} & \textbf{34.95} \\
    \bottomrule
  \end{tabular}
\end{table*}
With the predicted occlusion mask, \knmrA{an alternative approach} to improve the rendering quality in occluded regions \knmrA{would be} to mask out the occluded regions on the rendered image and inpaint. However, we observe that \knmrA{such} inpainting introduces two issues. Firstly, the occluded regions do not necessarily contain artifacts.
\wA{Fig.~\ref{fig:inp_or_ref} shows that,}
while part of the patterns (highlighted in ``GT occlusion'') on the pants are not observable from the input views, the input views provide similar patterns that can be used to hallucinate the occluded regions even without using image space \wA{refinement} (shown in ``w/o refinement''). Masking out the occluded regions in the case described above destroys available information for hallucination. Secondly, it can be challenging for the inpainting network to generate complex patterns (shown in ``w/ inpainting'') unless using a large model capacity, which \knmrA{would slow} down the rendering process. Therefore, we argue that the refinement-based method is more flexible in keeping the available information (shown in ``w/ refinement'') and hallucinating only when necessary, suiting our task that requires efficient rendering.
\begin{figure*}[ht]
  \centering
  \includegraphics[width=1\linewidth]{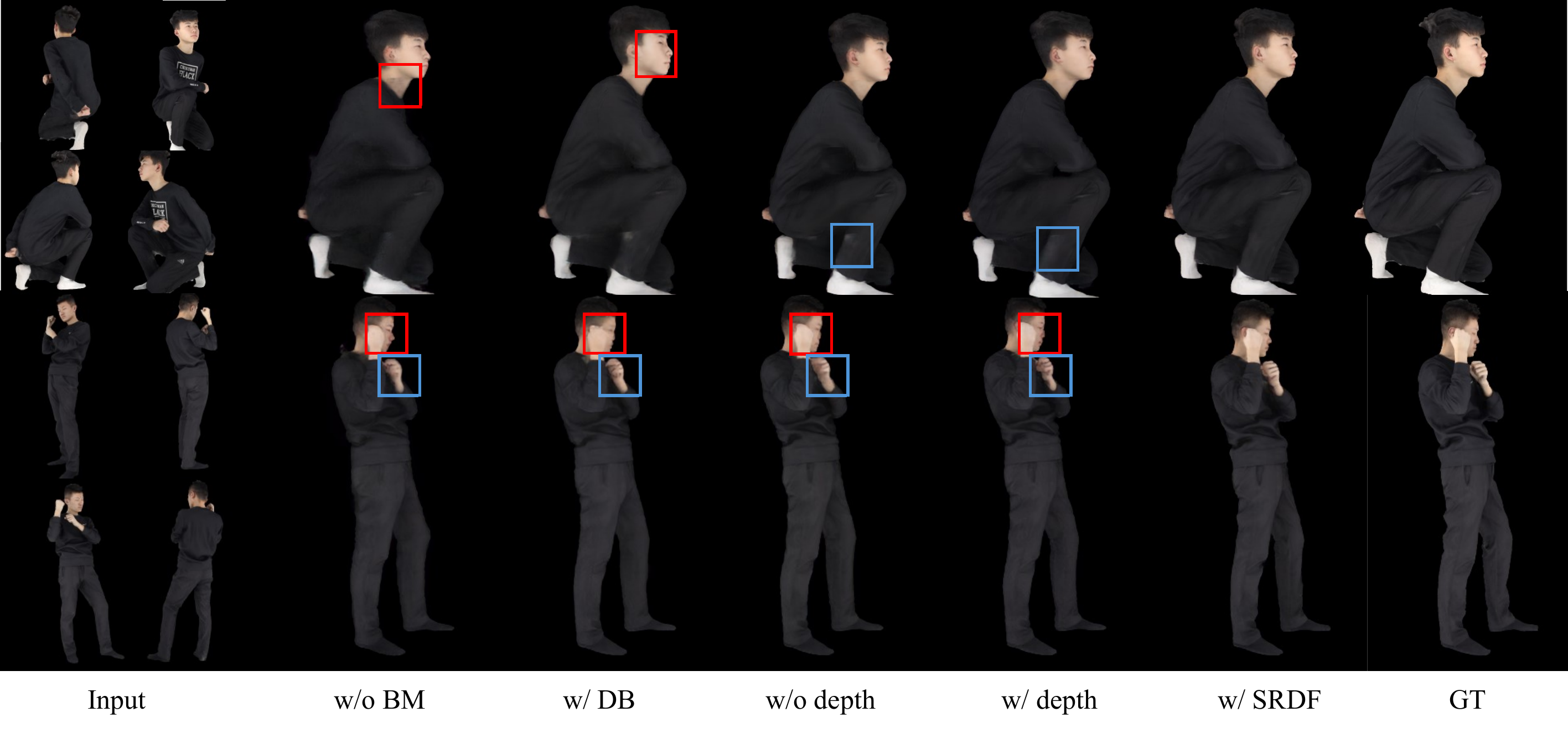}
  \caption{Qualitative results of the ablation studies. ``w/o BM'' denotes the results without using the boundary meshes. ``w/ DB'' denotes the results with the density-based rendering. ``w/o depth'' denotes the results without using the depth loss. ``w/ depth'' denotes the results with the depth loss only. ``w/ SRDF'' denotes the results with the depth loss and the SRDF loss.
  \wA{As shown in the red boxes in the first row, results from the model missing components exhibit artifacts on neck or unnatural face. The results from the model without the SRDF loss exhibit slight artifacts on the pants, as shown in the blue boxes in the first row. In the second row, models except for the one using SRDF loss exhibit different extents of artifacts on face and one hand.}}
  \label{fig:ablation1}
\end{figure*}
\subsection{Loss Functions} \label{sec:loss}
We train our network in an end-to-end manner using a loss function that includes an color loss \(\mathcal{L}_\mathit{color}\) and a perceptual loss \(\mathcal{L}_\mathit{percep}\). When using ground\knmrA{-}truth depth, we also include a depth loss \(\knmrA{\mathcal{L}}_\mathit{depth}\), and an SRDF loss \(\mathcal{L}_\mathit{SRDF}\). The total loss when using the ground\knmrA{-}truth depth is defined as follows:
\begin{equation}
  \begin{split}
    \mathcal{L} = \lambda_\mathit{color} \, \mathcal{L}_\mathit{color} + \lambda_\mathit{percep} \, \mathcal{L}_\mathit{percep}\\
    + \lambda_\mathit{depth} \, \mathcal{L}_\mathit{depth} + \lambda_\mathit{SRDF} \, \mathcal{L}_\mathit{SRDF},
  \end{split}
\end{equation}
\knmrA{where $\lambda_*$ are weights of the corresponding loss functions.}
The color loss is defined as the \(\ell_1\) loss between the predicted colors and the ground\knmrA{-}truth colors on pixels. The perceptual loss is defined as the \(\ell_1\) loss between the VGG~\cite{vgg} features of the predicted and the ground\knmrA{-}truth images. Both the color and perceptual loss are commonly used in NeRFs for humans~\cite{keypointnerf, gmnerf}. When the input views are as sparse as three, we fine-tune the model trained with \(\mathcal{L}\) with loss \(\mathcal{L} + \lambda_\mathit{adv} \mathcal{L}_\mathit{adv}\) additionally using a discriminator \(\mathbf{D}\).
\(\mathcal{L}_\mathit{adv}\) includes a non-saturating adversarial loss, a gradient penalty~\cite{gp}, and a feature matching loss~\cite{highres}\knmrA{,} as done in LaMa~\cite{lama}. We stop the gradient backpropagation for the non-occluded regions in the adversarial loss to avoid hallucination in observable regions.

While our method can generalize to novel subjects without training with ground\knmrA{-}truth depth (by setting \(\lambda_\mathit{depth} = 0\) and \(\lambda_\mathit{SRDF} = 0\)), we can further improve the rendering quality and geometry accuracy when the ground\knmrA{-}truth depth is available. We use the depth loss \(\mathcal{L}_\mathit{depth}\) as the \(\ell_1\)-distance between the predicted depth map \(\hat{I}_d\) and the ground\knmrA{-}truth depth map. Thanks to the SRDF-based rendering, we can easily calculate the ground\knmrA{-}truth SRDF values for each sample position from the ground\knmrA{-}truth depth map and employ the SRDF loss \(\mathcal{L}_\mathit{SRDF}\) to enforce the SRDFs representing actual surfaces. The SRDF loss \(\mathcal{L}_\mathit{SRDF}\) is defined as the \(\ell_1\)-distance between the predicted SRDF values \(f\) and the ground\knmrA{-}truth SRDF values \(f^{gt}\) at the two sets of ray sampling positions \(\{t_k\}_{k=1}^K\) and \(\{t_l\}_{l=1}^L\) used for sampling guidance and rendering, respectively. Formally, the SRDF loss is defined as follows:
\begin{equation}
  \begin{split}
    \mathcal{L}_\mathit{SRDF} \! = \! \frac{1}{K} \! \sum_{k=1}^K \left|f(t_k) \! - \! f^{gt}(t_k)\right| \! + \! \frac{1}{L} \! \sum_{l=1}^L \left|f(t_l) \! - \! f^{gt}(t_l)\right|.
  \end{split}
\end{equation}
To calculate the ground\knmrA{-}truth SRDF values \(f^{gt}(t)\) at \(t\), we first calculate the \wA{ray parameters} \(\{t_m\}_{m=1}^M\) of intersection points between the ray and the ground\knmrA{-}truth human scan with PyTorch3D~\cite{pytorch3d}, and then calculate \(f^{gt}(t)\) as follows:
\begin{equation}
  \begin{split}
    t_{\mathit{closest}} = \argmin_{m=1}^M \left|t - t_m\right|, \\
    f^{\mathit{gt}}(t) = t_{\mathit{closest}} - t,
  \end{split}
\end{equation}
where \(t_{closest}\) is the \wA{ray parameter} of the closest intersection point to the sample at \(t\). We observe that the SRDF loss helps to further improve the rendering quality,
which will be shown in Sec.~\ref{sec:ablation}.
\section{Experiments} \label{sec:exp}
In this section, we show the evaluation of our method for novel view synthesis. We first introduce our experimental settings, including datasets and implementation details in Sec.~\ref{sec:exp_settings}. We compared our method with the state-of-the-art methods on the novel view synthesis task, with results shown in Sec.~\ref{sec:comp_qual}. We also conducted ablation studies to analyze the effectiveness of our method, which is shown in Sec.~\ref{sec:ablation}.
\subsection{Experimental Settings} \label{sec:exp_settings}
We evaluated our method on two datasets: ZJU-MoCap~\cite{neuralbody} and THuman2.0~\cite{thu2}. We introduce the datasets and the evaluation protocol in Sec.~\ref{sec:dataset} and provide the implementation details in Sec.~\ref{sec:impl}.
\subsubsection{Datasets and Evaluation Protocols} \label{sec:dataset}
ZJU-MoCap~\cite{neuralbody} is a real-world multi-view human dataset that contains ten subjects performing various actions. Following the evaluation protocol adopted in KeyPointNeRF~\cite{keypointnerf} and GM-NeRF~\cite{gmnerf}, we split the dataset into seven subjects for training and three subjects for testing using three input views. THuman2.0~\cite{thu2} is a synthetic human dataset that contains 525 human scan meshes. To render the multi-view images for training, we placed the camera at a distance so that the human height fits the image height and sample a roll angle uniformly within \([0^\circ, 360^\circ]\) with an interval of \(10^\circ\). We split the dataset into 400 meshes for training and 125 meshes for testing using four input views with an interval of \(90^\circ\), aligning with the input view number in GM-NeRF~\cite{gmnerf}. The angle between target and input views is \(45^\circ\). The resolution of the input images is \(1024 \times 1024\), and the resolution of the rendered images is \(512 \times 512\).
For evaluation metrics, we adopted peak signal-to-noise ratio (PSNR), structural similarity index (SSIM)~\cite{ssim}, and learned perceptual image patch similarity (LPIPS)~\cite{lpips}, which are commonly used in the novel view synthesis task. We used a single NVIDIA RTX A6000 GPU for rendering speed evaluation.

\begin{table}[ht]
  \centering
  \caption{Ablation studies on the THuman2.0 dataset using four input views. Either removing the boundary meshes or using the density-based rendering reduces the rendering quality. Using the depth loss and the SRDF loss further improves the rendering quality.}
  \label{tab:abl1}
  \resizebox{0.49\textwidth}{!}{%
  \begin{tabular}{l|ccc}
      \toprule
      Method & PSNR \(\uparrow\) & SSIM \(\uparrow\) & LPIPS \(\downarrow\) \\
      \midrule
      Ours (w/o boundary meshes) & 29.08 & 0.950 & 0.068 \\
      Ours (w/ density-based) & 28.69 & 0.934 & 0.075 \\
      Ours (w/o depth) & \textbf{29.44} & \textbf{0.951} & \textbf{0.060} \\
      Ours (w/ depth only) & 29.60 & 0.968 & 0.051 \\
      Ours (w/ depth \& SRDF) & \textbf{31.73} & \textbf{0.973} & \textbf{0.034} \\
      \bottomrule
  \end{tabular}
  }
\end{table}

\begin{table}[ht]
  \centering
  \caption{Ablation studies on the THuman2.0 dataset to analyze the effectiveness of handling occluded regions. We use three input views to obtain the results. ``Ours w/o refinement'' denotes the results without using image space \wA{refinement}. ``Ours w/ occlusion-awareness'' denotes the results with image space refinement using occlusion-aware features.}
  \label{tab:abl2}
  \resizebox{0.49\textwidth}{!}{%
  \begin{tabular}{l|ccc}
      \toprule
      Method & PSNR \(\uparrow\) & SSIM \(\uparrow\) & LPIPS \(\downarrow\) \\
      \midrule
      Ours w/o refinement & 30.57 & 0.965 & 0.048 \\
      Ours w/ occlusion-awareness & \textbf{31.03} & \textbf{0.970} & \textbf{0.035} \\
      \bottomrule
  \end{tabular}
  }
\end{table}

\subsubsection{Implementation Details} \label{sec:impl}
We implemented our method using PyTorch. We used Adam optimizer with a learning rate of \(10^{-4}\) and a batch size of one image to train the network until convergence. We set the number\knmrA{s} of samples per ray
\knmrA{$K = 16$}
for sampling guidance and
\knmrA{$K = 8$}
for \knmrA{volume} rendering. We set the resolution of the geometry feature volume to \(224 \times 224 \times 224\). For the loss weights,
we set \(\lambda_\mathit{color} \knmrA{=} 150\), \(\lambda_\mathit{percep} \knmrA{=} 0.5\), \(\lambda_\mathit{adv} \knmrA{=} 5\), \(\lambda_\mathit{depth} \knmrA{=} 1\), and \(\lambda_\mathit{SRDF} \knmrA{=} 1\).
Please refer to Appendix~\ref{sec:impl_add} for more implementation details.
\begin{figure*}[ht]
  \centering
  \includegraphics[width=1\linewidth]{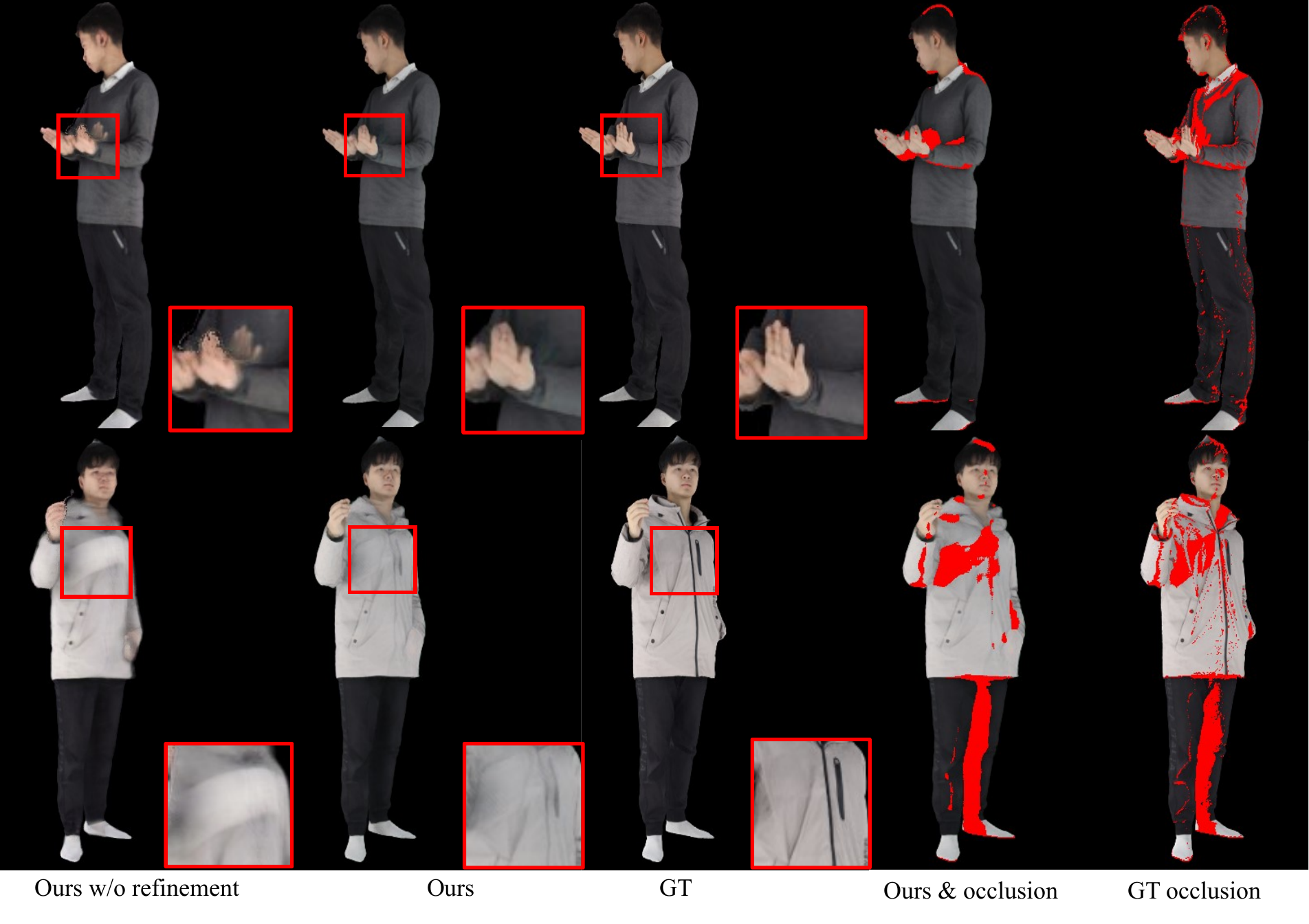}
  \caption{Qualitative results of the ablation studies on handling occluded regions. We use three input views of THuman2.0 dataset to obtain the results. ``Ours w/o refinement'' denotes the results without using image space \wA{refinement}. ``Ours'' denotes the results with image space refinement using occlusion-aware features. ``Ours \& occlusion'' denotes the rendered results with the predicted occlusion mask highlighted in red. ``GT occlusion'' denotes the ground\knmrA{-}truth occlusion mask.}
  \label{fig:ablation2}
\end{figure*}
\begin{figure*}[ht]
  \centering
  \includegraphics[width=0.95\linewidth]{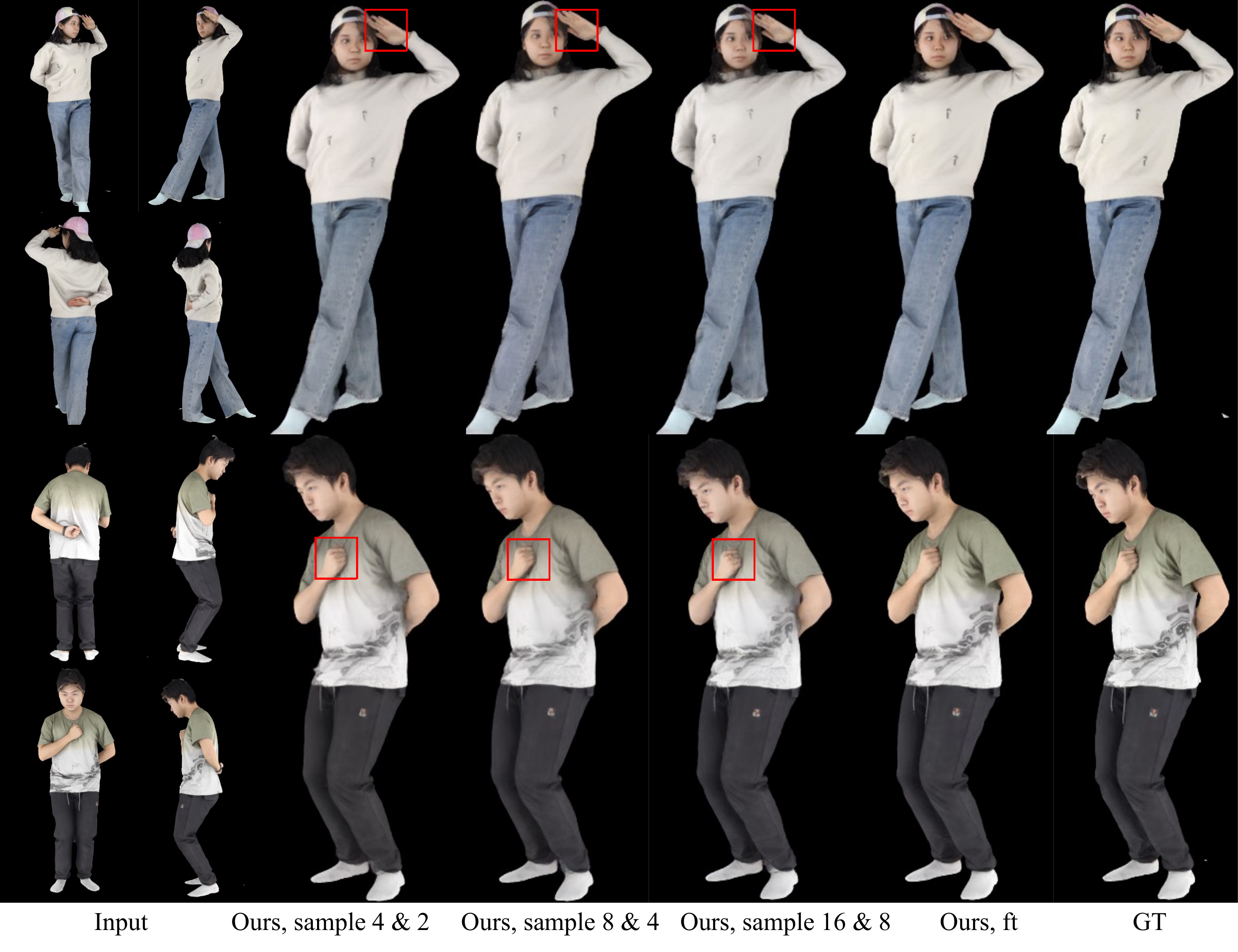}
  \caption{Qualitative results of the ablation studies of the number of samples per ray and the training protocol. ``Ours, 4 \& 2'' denotes our method trained with four samples per ray for sampling guidance and two samples per ray for rendering. ``Ours, ft'' denotes our method fine-tuned using images of target subjects.
  \wA{
  While the results from the model using different numbers of samples per ray exhibit similar quality, their rendering of human hands, as shown in the red boxes, are less accurate than the results from the fine-tuned model.}}
  \label{fig:ablation3}
\end{figure*}

\subsection{Comparisons on Novel View Synthesis} \label{sec:comp_qual}
We compared our method with the state-of-the-art generalizable methods that prioritize rendering quality on novel subjects, including IBRNet~\cite{ibrnet}, KeypointNeRF~\cite{keypointnerf}, and GM-NeRF~\cite{gmnerf}. We also compared with the speed-prioritized NeRF-based methods, including GP-NeRF~\cite{gpnerf}, ENeRF~\cite{enerf}. We retrained and evaluated the networks of the compared methods on ZJU-MoCap~\cite{neuralbody} and THuman2.0 dataset~\cite{thu2} using the dataset split and evaluation protocol introduced in Sec.~\ref{sec:dataset}. We implemented GM-NeRF following the description in the paper. For other methods, we used the official implementation and training protocols.
Table~\ref{tab:quality} shows the quantitative comparison results.
We observe that our method outperforms the state-of-the-art methods in terms of rendering quality 
\wA{when the ground\knmrA{-}truth depths are not available.}
When using the ground\knmrA{-}truth depth, rendering quality of our method is further improved.
\wA{Fig.~\ref{fig:eval_qual} shows the qualitative results of our method compared with the state-of-the-art methods. We observe that our method can synthesize high-quality novel views with much less artifacts, blurriness, and better details compared to the state-of-the-art methods. Please refer to Appendix~\ref{sec:exp_add} for more qualitative results.}
We also compared with the state-of-the-art 3D-Gaussian-based method for human, GPS-Gaussian~\cite{gps}. GPS-Gaussian requires at least six inputs and the ground\knmrA{-}truth depth to synthesize complete human bodies due to its usage of stereo-based depth estimation. We used six input views and ground\knmrA{-}truth depth for GPS-Gaussian to obtain the results shown in Table~\ref{tab:quality}. Our method still outperforms GPS-Gaussian even with much fewer input views and without ground\knmrA{-}truth depth.

\begin{table}[ht]
  \centering
  \caption{Ablation studies of the number of samples per ray on the THuman2.0 dataset. ``4 \& 2'' denotes the number of samples per ray for sampling guidance and rendering, respectively.}
  \label{tab:abl_sample}
  \begin{tabular}{l|ccc}
      \toprule
      Samples & PSNR \(\uparrow\) & SSIM \(\uparrow\) & LPIPS \(\downarrow\) \\
      \midrule
      4 \& 2 & 31.47 & 0.971 & 0.036 \\
      8 \& 4 & 31.50 & 0.971 & 0.035 \\
      16 \& 8 & \textbf{31.73} & \textbf{0.973} & \textbf{0.034} \\
      \bottomrule
  \end{tabular}
\end{table}

\begin{table}[ht]
  \centering
  \caption{Ablation studies of the number of input views on the THuman2.0 dataset. ``Interval'' denotes the angle between input views.}
  \label{tab:abl_view}
  \begin{tabular}{l|ccc}
      \toprule
      Views (Interval) & PSNR \(\uparrow\) & SSIM \(\uparrow\) & LPIPS \(\downarrow\) \\
      \midrule
      3 (\(180^\circ\)) & 31.03 & 0.970 & 0.035 \\
      4 (\(90^\circ\)) & 31.73 & 0.973 & 0.034\\
      5 (\(70^\circ\)) & 34.48 & 0.985 & 0.023 \\
      6 (\(60^\circ\)) & \textbf{34.77} & \textbf{0.986} & \textbf{0.021} \\
      \bottomrule
  \end{tabular}
\end{table}

\begin{table}[ht]
  \centering
  \caption{Ablation study of fine-tuning the model using images of target subjects on the THuman2.0 dataset. We report the averaged metrics on 125 test subjects for both zero-shot and fine-tuned models.}
  \label{tab:abl_finetune}
  \begin{tabular}{l|ccc}
      \toprule
      Samples & PSNR \(\uparrow\) & SSIM \(\uparrow\) & LPIPS \(\downarrow\) \\
      \midrule
      Ours (zero-shot) & 31.73 & 0.973 & 0.034 \\
      Ours (fine-tuned) & \textbf{36.29} & \textbf{0.988} & \textbf{0.018} \\
      \bottomrule
  \end{tabular}
\end{table}

\subsection{Ablation Studies} \label{sec:ablation}
We conducted ablation studies on the components of our method to analyze their effectiveness (Sec.~\ref{sec:abl_comp}) and the experimental settings to analyze their impact on the rendering quality (Sec.~\ref{sec:abl_exp}).
\subsubsection{Studies on Method Components} \label{sec:abl_comp}
We conducted ablation studies to analyze the effectiveness of our method using the THuman2.0 dataset. For the setting without ground\knmrA{-}truth depth, we analyzed the effect of using the boundary meshes by removing the sampling guidance of the boundary meshes when sampling \(\{\mathbf{p}_k\}_{k=1}^K\). Since the boundary meshes is removed, we also removed the usage of the feature \(\mathbf{g}_\mathit{stat}\) on them.
\wA{We also analyzed the usage of the SRDF formulation for rendering by replacing the SRDF formulation with the density formulation used in NeRF~\cite{nerf}.}
\wA{Both the quantitative results in Table~\ref{tab:abl1} and the qualitative results in Fig.~\ref{fig:ablation1}}
show that removing the boundary meshes or
\wA{altering the SRDF formulation}
degrades the rendering quality.

In the setting using ground\knmrA{-}truth depth, we analyzed the effect of the SRDF loss by removing it and only using the depth loss.
\wA{The results in Table~\ref{tab:abl1} and Fig.~\ref{fig:ablation1} show that}
the SRDF loss further improves the rendering quality.

We also analyzed the effectiveness of our method in handling occluded regions on the THuman2.0 dataset. We used a setting of three input views, which is more likely to contain occluded regions, along with using ground\knmrA{-}truth geometry. We compared the results without using the image space refinement with the results using the image space refinement and occlusion-aware features. The results in Table~\ref{tab:abl2} and Fig.~\ref{fig:ablation2} show that the image space refinement improves the rendering quality by removing artifacts in occluded regions. We also visualized the predicted occlusion mask and the ground\knmrA{-}truth occlusion mask, demonstrating that occlusion-aware features effectively guide the network to regress the occluded regions.

\subsubsection{Studies on Experimental Settings} \label{sec:abl_exp}
In addition to ablation studies on the proposed components, we also conducted ablation studies on the number of samples per ray, the number of input views, and the training protocol. The three ablation studies are conducted on the THuman2.0 dataset using four input views and ground\knmrA{-}truth geometry.
\knmrA{Tables}~\ref{tab:abl_sample} and \ref{tab:abl_view}
show the quantitative results of the first two studies, respectively.
We observed that our method does not suffer significantly from reducing the number of samples per ray to a quarter of the original number, namely, \(K=4\) and \(L=2\). The insensitive rendering quality to the number of samples per ray indicates that our method can potentially handle high-resolution images at a similar rendering speed by further reducing the number of samples. Given a view number \(N_I\), we chose the camera poses in the same way described in Sec.~\ref{sec:dataset}, so that a smaller input number results in a larger view interval of \(\frac{360^\circ}{N_I}\).

We also conducted an ablation study on the training protocol by fine-tuning the model using images of target subjects. Specifically, we use 32 images of target subjects \wA{to fine-tune all trainable parameters of the model.}
Table~\ref{tab:abl_finetune} shows the averaged quantitative results on 125 test subjects from the THuman2.0 dataset.
We observed that fine-tuning the model using images of target subjects further improves the rendering quality. 
Fig.~\ref{fig:ablation3} shows some qualitative results on experimental settings.
\wA{Please refer to Appendix~\ref{sec:exp_add} for more ablation study results.}
\section{Conclusion and Future Work}
\label{sec:conclusion}

We \knmrA{have presented EG-HumanNeRF,} an efficient and generalizable human NeRF method for sparse inputs. Our method outperforms existing quality-prioritized methods in rendering quality and achieves competitive rendering speed compared with speed-prioritized methods. Utilizing the SMPL mesh as a sampling restriction allows our method to render high-quality images with a small number of samples. Without introducing heavyweight generative models, our method improves the rendering quality in occluded regions through occlusion-aware attention and image space refinement.
\wA{SRDF formulation}
and the SRDF loss further improve the rendering quality.

\subsection{Limitations and Future Work}
Our work shares limitations with the optimization-based method utilizing boundary meshes~\cite{adashell}: there may be missing geometries if the boundary meshes does not cover the entire human body, e.g., for humans wearing loose clothes. A possible solution to this issue is to partially loosen the boundary meshes' restriction at the fine-tuning stage.

An interesting direction for future work is to explore the capability of our method handling high-resolution image rendering. To avoid the rendering time increase with the image resolution, we can simply reduce \knmrA{the} number of sample\knmrA{s} per ray, as the output image quality with our method is insensitive to the number of samples per ray. Improving the architecture of the 2D CNN \knmrA{$\mathbf{U}$} for efficient and upsampling to high-resolution images is also a promising direction.

{\small
\bibliographystyle{ieeenat_fullname}
\bibliography{11_references}

\begin{thebibliography}{63}
\providecommand{\natexlab}[1]{#1}
\providecommand{\url}[1]{\texttt{#1}}
\expandafter\ifx\csname urlstyle\endcsname\relax
  \providecommand{\doi}[1]{doi: #1}\else
  \providecommand{\doi}{doi: \begingroup \urlstyle{rm}\Url}\fi

\bibitem[Barron et~al.(2021)Barron, Mildenhall, Tancik, Hedman, Martin{-}Brualla, and Srinivasan]{mipnerf}
Jonathan~T. Barron, Ben Mildenhall, Matthew Tancik, Peter Hedman, Ricardo Martin{-}Brualla, and Pratul~P. Srinivasan.
\newblock Mip-nerf: {A} multiscale representation for anti-aliasing neural radiance fields.
\newblock In \emph{ICCV}, pages 5835--5844, 2021.

\bibitem[Cao et~al.(2023)Cao, Wang, Chemerys, Shakhrai, Hu, Fu, Makoviichuk, Tulyakov, and Ren]{mobiler2l}
Junli Cao, Huan Wang, Pavlo Chemerys, Vladislav Shakhrai, Ju Hu, Yun Fu, Denys Makoviichuk, Sergey Tulyakov, and Jian Ren.
\newblock Real-time neural light field on mobile devices.
\newblock In \emph{CVPR}, pages 8328--8337, 2023.

\bibitem[Chan et~al.(2022)Chan, Lin, Chan, Nagano, Pan, Mello, Gallo, Guibas, Tremblay, Khamis, Karras, and Wetzstein]{eg3d}
Eric~R. Chan, Connor~Z. Lin, Matthew~A. Chan, Koki Nagano, Boxiao Pan, Shalini~De Mello, Orazio Gallo, Leonidas~J. Guibas, Jonathan Tremblay, Sameh Khamis, Tero Karras, and Gordon Wetzstein.
\newblock Efficient geometry-aware 3d generative adversarial networks.
\newblock In \emph{CVPR}, pages 16102--16112, 2022.

\bibitem[Chen et~al.(2021{\natexlab{a}})Chen, Xu, Zhao, Zhang, Xiang, Yu, and Su]{mvsnerf}
Anpei Chen, Zexiang Xu, Fuqiang Zhao, Xiaoshuai Zhang, Fanbo Xiang, Jingyi Yu, and Hao Su.
\newblock Mvsnerf: Fast generalizable radiance field reconstruction from multi-view stereo.
\newblock In \emph{ICCV}, pages 14104--14113, 2021{\natexlab{a}}.

\bibitem[Chen et~al.(2022{\natexlab{a}})Chen, Xu, Geiger, Yu, and Su]{tenso}
Anpei Chen, Zexiang Xu, Andreas Geiger, Jingyi Yu, and Hao Su.
\newblock Tensorf: Tensorial radiance fields.
\newblock In \emph{ECCV}, pages 333--350, 2022{\natexlab{a}}.

\bibitem[Chen et~al.(2021{\natexlab{b}})Chen, Zhang, Kang, Zhe, Bao, Jia, and Lu]{animnerf}
Jianchuan Chen, Ying Zhang, Di Kang, Xuefei Zhe, Linchao Bao, Xu Jia, and Huchuan Lu.
\newblock Animatable neural radiance fields from monocular rgb videos, 2021{\natexlab{b}}.

\bibitem[Chen et~al.(2023{\natexlab{a}})Chen, Yi, Ma, Jia, and Lu]{gmnerf}
Jianchuan Chen, Wentao Yi, Liqian Ma, Xu Jia, and Huchuan Lu.
\newblock Gm-nerf: Learning generalizable model-based neural radiance fields from multi-view images.
\newblock In \emph{CVPR}, pages 20648--20658, 2023{\natexlab{a}}.

\bibitem[Chen et~al.(2022{\natexlab{b}})Chen, Zhang, Xu, Liu, Cai, Feng, and Yan]{gpnerf}
Mingfei Chen, Jianfeng Zhang, Xiangyu Xu, Lijuan Liu, Yujun Cai, Jiashi Feng, and Shuicheng Yan.
\newblock Geometry-guided progressive nerf for generalizable and efficient neural human rendering.
\newblock In \emph{ECCV}, pages 222--239, 2022{\natexlab{b}}.

\bibitem[Chen et~al.(2023{\natexlab{b}})Chen, Funkhouser, Hedman, and Tagliasacchi]{mobilenerf}
Zhiqin Chen, Thomas~A. Funkhouser, Peter Hedman, and Andrea Tagliasacchi.
\newblock Mobilenerf: Exploiting the polygon rasterization pipeline for efficient neural field rendering on mobile architectures.
\newblock In \emph{CVPR}, pages 16569--16578, 2023{\natexlab{b}}.

\bibitem[Cheng et~al.(2022)Cheng, Xu, Piao, Qian, Wu, Lin, and Li]{gnp}
Wei Cheng, Su Xu, Jingtan Piao, Chen Qian, Wayne Wu, Kwan{-}Yee Lin, and Hongsheng Li.
\newblock Generalizable neural performer: Learning robust radiance fields for human novel view synthesis.
\newblock \emph{CoRR}, abs/2204.11798, 2022.

\bibitem[Dou et~al.(2016)Dou, Khamis, Degtyarev, Davidson, Fanello, Kowdle, Orts{-}Escolano, Rhemann, Kim, Taylor, Kohli, Tankovich, and Izadi]{depth_multi}
Mingsong Dou, Sameh Khamis, Yury Degtyarev, Philip Davidson, Sean~Ryan Fanello, Adarsh Kowdle, Sergio Orts{-}Escolano, Christoph Rhemann, David Kim, Jonathan Taylor, Pushmeet Kohli, Vladimir Tankovich, and Shahram Izadi.
\newblock Fusion4d: real-time performance capture of challenging scenes.
\newblock \emph{{ACM} Trans. Graph.}, 35\penalty0 (4):\penalty0 114:1--114:13, 2016.

\bibitem[Graham et~al.(2018)Graham, Engelcke, and van~der Maaten]{sparseconvnet}
Benjamin Graham, Martin Engelcke, and Laurens van~der Maaten.
\newblock 3d semantic segmentation with submanifold sparse convolutional networks.
\newblock In \emph{CVPR}, pages 9224--9232, 2018.

\bibitem[Gu et~al.(2022)Gu, Liu, Wang, and Theobalt]{stylenerf}
Jiatao Gu, Lingjie Liu, Peng Wang, and Christian Theobalt.
\newblock Stylenerf: {A} style-based 3d aware generator for high-resolution image synthesis.
\newblock In \emph{The Tenth International Conference on Learning Representations, {ICLR}}, pages 1--25, 2022.

\bibitem[Guo et~al.(2019)Guo, Lincoln, Davidson, Busch, Yu, Whalen, Harvey, Orts{-}Escolano, Pandey, Dourgarian, Tang, Tkach, Kowdle, Cooper, Dou, Fanello, Fyffe, Rhemann, Taylor, Debevec, and Izadi]{densecam_volume}
Kaiwen Guo, Peter Lincoln, Philip Davidson, Jay Busch, Xueming Yu, Matt Whalen, Geoff Harvey, Sergio Orts{-}Escolano, Rohit Pandey, Jason Dourgarian, Danhang Tang, Anastasia Tkach, Adarsh Kowdle, Emily Cooper, Mingsong Dou, Sean~Ryan Fanello, Graham Fyffe, Christoph Rhemann, Jonathan Taylor, Paul~E. Debevec, and Shahram Izadi.
\newblock The relightables: volumetric performance capture of humans with realistic relighting.
\newblock \emph{{ACM} Trans. Graph.}, 38\penalty0 (6):\penalty0 217:1--217:19, 2019.

\bibitem[Hedman et~al.(2021)Hedman, Srinivasan, Mildenhall, Barron, and Debevec]{snerg}
Peter Hedman, Pratul~P. Srinivasan, Ben Mildenhall, Jonathan~T. Barron, and Paul Debevec.
\newblock Baking neural radiance fields for real-time view synthesis.
\newblock \emph{ICCV}, 2021.

\bibitem[Hong et~al.(2023)Hong, Yu, Dai, Yang, Lian, Liu, Xu, Dong, and Wang]{torchsparse}
Ke Hong, Zhongming Yu, Guohao Dai, Xinhao Yang, Yaoxiu Lian, Zehao Liu, Ningyi Xu, Yuhan Dong, and Yu Wang.
\newblock Exploiting hardware utilization and adaptive dataflow for efficient sparse convolution in 3d point clouds.
\newblock In \emph{Proceedings of the Sixth Conference on Machine Learning and Systems, MLSys}, pages 428--441, 2023.

\bibitem[Hu et~al.(2022)Hu, Liu, Chen, Shen, and Jia]{efficient}
Tao Hu, Shu Liu, Yilun Chen, Tiancheng Shen, and Jiaya Jia.
\newblock Efficientnerf - efficient neural radiance fields.
\newblock In \emph{CVPR}, pages 12892--12901, 2022.

\bibitem[Jain et~al.(2021)Jain, Tancik, and Abbeel]{dietnerf}
Ajay Jain, Matthew Tancik, and Pieter Abbeel.
\newblock Putting nerf on a diet: Semantically consistent few-shot view synthesis.
\newblock In \emph{{ICCV} 2021}, pages 5865--5874, 2021.

\bibitem[Kanaoka et~al.(2023)Kanaoka, Sonogashira, Tamukoh, and Kawanishi]{manifoldnerf}
Daiju Kanaoka, Motoharu Sonogashira, Hakaru Tamukoh, and Yasutomo Kawanishi.
\newblock Manifoldnerf: View-dependent image feature supervision for few-shot neural radiance fields.
\newblock In \emph{34th British Machine Vision Conference 2023, {BMVC} 2023}, pages 1--13, 2023.

\bibitem[Kim et~al.(2022)Kim, Seo, and Han]{infonerf}
Mijeong Kim, Seonguk Seo, and Bohyung Han.
\newblock Infonerf: Ray entropy minimization for few-shot neural volume rendering.
\newblock In \emph{{CVPR} 2022}, pages 12902--12911, 2022.

\bibitem[Kurz et~al.(2022)Kurz, Neff, Lv, Zollh{\"{o}}fer, and Steinberger]{adanerf}
Andreas Kurz, Thomas Neff, Zhaoyang Lv, Michael Zollh{\"{o}}fer, and Markus Steinberger.
\newblock Adanerf: Adaptive sampling for real-time rendering of neural radiance fields.
\newblock In \emph{ECCV}, pages 254--270, 2022.

\bibitem[Kwon et~al.(2021)Kwon, Kim, Ceylan, and Fuchs]{nhp}
Youngjoong Kwon, Dahun Kim, Duygu Ceylan, and Henry Fuchs.
\newblock Neural human performer: Learning generalizable radiance fields for human performance rendering.
\newblock In \emph{NeurIPS}, pages 24741--24752, 2021.

\bibitem[Kwon et~al.(2024)Kwon, Fang, Lu, Dong, Zhang, Carrasco, Mosella-Montoro, Xu, Takagi, Kim, Prakash, and la~Torre]{ghg}
Youngjoong Kwon, Baole Fang, Yixing Lu, Haoye Dong, Cheng Zhang, Francisco~Vicente Carrasco, Albert Mosella-Montoro, Jianjin Xu, Shingo Takagi, Daeil Kim, Aayush Prakash, and Fernando~De la Torre.
\newblock Generalizable human gaussians for sparse view synthesis, 2024.

\bibitem[Li et~al.(2024)Li, Gou, and Tan]{idnerf}
Yaokun Li, Chao Gou, and Guang Tan.
\newblock Id-nerf: Indirect diffusion-guided neural radiance fields for generalizable view synthesis, 2024.

\bibitem[Li et~al.(2022)Li, Yu, Zheng, and Liu]{depth_single_22}
Zhe Li, Tao Yu, Zerong Zheng, and Yebin Liu.
\newblock Robust and accurate 3d self-portraits in seconds.
\newblock \emph{{IEEE} Trans. Pattern Anal. Mach. Intell.}, 44\penalty0 (11):\penalty0 7854--7870, 2022.

\bibitem[Lin et~al.(2022)Lin, Peng, Xu, Yan, Shuai, Bao, and Zhou]{enerf}
Haotong Lin, Sida Peng, Zhen Xu, Yunzhi Yan, Qing Shuai, Hujun Bao, and Xiaowei Zhou.
\newblock Efficient neural radiance fields for interactive free-viewpoint video.
\newblock In \emph{{SIGGRAPH} Asia 2022 Conference Papers}, pages 39:1--39:9, 2022.

\bibitem[Liu et~al.(2020)Liu, Gu, Lin, Chua, and Theobalt]{voxel}
Lingjie Liu, Jiatao Gu, Kyaw~Zaw Lin, Tat{-}Seng Chua, and Christian Theobalt.
\newblock Neural sparse voxel fields.
\newblock In \emph{NeurIPS}, pages 15651--15663, 2020.

\bibitem[Loper et~al.(2015)Loper, Mahmood, Romero, Pons{-}Moll, and Black]{smpl}
Matthew Loper, Naureen Mahmood, Javier Romero, Gerard Pons{-}Moll, and Michael~J. Black.
\newblock {SMPL:} a skinned multi-person linear model.
\newblock \emph{{ACM} Trans. Graph.}, 34\penalty0 (6):\penalty0 248:1--248:16, 2015.

\bibitem[Mihajlovic et~al.(2022)Mihajlovic, Bansal, Zollh{\"{o}}fer, Tang, and Saito]{keypointnerf}
Marko Mihajlovic, Aayush Bansal, Michael Zollh{\"{o}}fer, Siyu Tang, and Shunsuke Saito.
\newblock Keypointnerf: Generalizing image-based volumetric avatars using relative spatial encoding of keypoints.
\newblock In \emph{ECCV}, pages 179--197, 2022.

\bibitem[Mildenhall et~al.(2020)Mildenhall, Srinivasan, Tancik, Barron, Ramamoorthi, and Ng]{nerf}
Ben Mildenhall, Pratul~P. Srinivasan, Matthew Tancik, Jonathan~T. Barron, Ravi Ramamoorthi, and Ren Ng.
\newblock Nerf: Representing scenes as neural radiance fields for view synthesis.
\newblock In \emph{ECCV}, pages 405--421, 2020.

\bibitem[M\"uller et~al.(2022)M\"uller, Evans, Schied, and Keller]{ngp}
Thomas M\"uller, Alex Evans, Christoph Schied, and Alexander Keller.
\newblock Instant neural graphics primitives with a multiresolution hash encoding.
\newblock \emph{ACM Trans. Graph.}, 41\penalty0 (4):\penalty0 102:1--102:15, 2022.

\bibitem[Neff et~al.(2021)Neff, Stadlbauer, Parger, Kurz, Mueller, Chaitanya, Kaplanyan, and Steinberger]{donerf}
Thomas Neff, Pascal Stadlbauer, Mathias Parger, Andreas Kurz, Joerg~H. Mueller, Chakravarty R.~Alla Chaitanya, Anton Kaplanyan, and Markus Steinberger.
\newblock Donerf: Towards real-time rendering of compact neural radiance fields using depth oracle networks.
\newblock \emph{Comput. Graph. Forum}, 40\penalty0 (4):\penalty0 45--59, 2021.

\bibitem[Pavlakos et~al.(2019)Pavlakos, Choutas, Ghorbani, Bolkart, Osman, Tzionas, and Black]{smplx}
Georgios Pavlakos, Vasileios Choutas, Nima Ghorbani, Timo Bolkart, Ahmed A.~A. Osman, Dimitrios Tzionas, and Michael~J. Black.
\newblock Expressive body capture: {3D} hands, face, and body from a single image.
\newblock In \emph{CVPR}, pages 10975--10985, 2019.

\bibitem[Peng et~al.(2021)Peng, Zhang, Xu, Wang, Shuai, Bao, and Zhou]{neuralbody}
Sida Peng, Yuanqing Zhang, Yinghao Xu, Qianqian Wang, Qing Shuai, Hujun Bao, and Xiaowei Zhou.
\newblock Neural body: Implicit neural representations with structured latent codes for novel view synthesis of dynamic humans.
\newblock In \emph{CVPR}, pages 9054--9063, 2021.

\bibitem[Ravi et~al.(2020)Ravi, Reizenstein, Novotny, Gordon, Lo, Johnson, and Gkioxari]{pytorch3d}
Nikhila Ravi, Jeremy Reizenstein, David Novotny, Taylor Gordon, Wan-Yen Lo, Justin Johnson, and Georgia Gkioxari.
\newblock Accelerating 3d deep learning with pytorch3d.
\newblock \emph{arXiv:2007.08501}, 2020.

\bibitem[Reiser et~al.(2023)Reiser, Szeliski, Verbin, Srinivasan, Mildenhall, Geiger, Barron, and Hedman]{merf}
Christian Reiser, Richard Szeliski, Dor Verbin, Pratul~P. Srinivasan, Ben Mildenhall, Andreas Geiger, Jonathan~T. Barron, and Peter Hedman.
\newblock {MERF:} memory-efficient radiance fields for real-time view synthesis in unbounded scenes.
\newblock \emph{{ACM} Trans. Graph.}, 42\penalty0 (4):\penalty0 89:1--89:12, 2023.

\bibitem[Ren et~al.(2023)Ren, Zhang, Pollefeys, S{\"u}sstrunk, and Wang]{volrecon}
Yufan Ren, Tong Zhang, Marc Pollefeys, Sabine S{\"u}sstrunk, and Fangjinhua Wang.
\newblock Volrecon: Volume rendering of signed ray distance functions for generalizable multi-view reconstruction.
\newblock In \emph{CVPR}, pages 16685--16695, 2023.

\bibitem[Ross and Doshi{-}Velez(2018)]{gp}
Andrew~Slavin Ross and Finale Doshi{-}Velez.
\newblock Improving the adversarial robustness and interpretability of deep neural networks by regularizing their input gradients.
\newblock In \emph{Proceedings of the Thirty-Second {AAAI} Conference on Artificial Intelligence}, pages 1660--1669, 2018.

\bibitem[Shao et~al.(2022)Shao, Zhang, Zhang, Chen, Cao, Yu, and Liu]{doublefield}
Ruizhi Shao, Hongwen Zhang, He Zhang, Mingjia Chen, Yanpei Cao, Tao Yu, and Yebin Liu.
\newblock Doublefield: Bridging the neural surface and radiance fields for high-fidelity human reconstruction and rendering.
\newblock In \emph{CVPR}, pages 15851--15861, 2022.

\bibitem[Simonyan and Zisserman(2015)]{vgg}
Karen Simonyan and Andrew Zisserman.
\newblock Very deep convolutional networks for large-scale image recognition.
\newblock In \emph{3rd International Conference on Learning Representations, {ICLR} 2015, San Diego, CA, USA, May 7-9, 2015, Conference Track Proceedings}, pages 1--14, 2015.

\bibitem[Sitzmann et~al.(2020)Sitzmann, Martel, Bergman, Lindell, and Wetzstein]{siren}
Vincent Sitzmann, Julien N.~P. Martel, Alexander~W. Bergman, David~B. Lindell, and Gordon Wetzstein.
\newblock Implicit neural representations with periodic activation functions.
\newblock In \emph{NeurIPS}, pages 7462--7473, 2020.

\bibitem[Su et~al.(2021)Su, Yu, Zollh{\"{o}}fer, and Rhodin]{anerf}
Shih{-}Yang Su, Frank Yu, Michael Zollh{\"{o}}fer, and Helge Rhodin.
\newblock A-nerf: Articulated neural radiance fields for learning human shape, appearance, and pose.
\newblock In \emph{NeurIPS}, pages 12278--12291, 2021.

\bibitem[Sun et~al.(2022)Sun, Sun, and Chen]{direct}
Cheng Sun, Min Sun, and Hwann{-}Tzong Chen.
\newblock Direct voxel grid optimization: Super-fast convergence for radiance fields reconstruction.
\newblock In \emph{CVPR}, pages 5449--5459, 2022.

\bibitem[Suvorov et~al.(2022)Suvorov, Logacheva, Mashikhin, Remizova, Ashukha, Silvestrov, Kong, Goka, Park, and Lempitsky]{lama}
Roman Suvorov, Elizaveta Logacheva, Anton Mashikhin, Anastasia Remizova, Arsenii Ashukha, Aleksei Silvestrov, Naejin Kong, Harshith Goka, Kiwoong Park, and Victor Lempitsky.
\newblock Resolution-robust large mask inpainting with fourier convolutions.
\newblock In \emph{WACV}, pages 3172--3182, 2022.

\bibitem[Tan and Le(2021)]{efficientnetv2}
Mingxing Tan and Quoc~V. Le.
\newblock Efficientnetv2: Smaller models and faster training.
\newblock In \emph{Proceedings of the 38th International Conference on Machine Learning, {ICML} 2021, 18-24 July 2021, Virtual Event}, pages 10096--10106, 2021.

\bibitem[Vlasic et~al.(2008)Vlasic, Baran, Matusik, and Popovic]{densecam_mesh_08}
Daniel Vlasic, Ilya Baran, Wojciech Matusik, and Jovan Popovic.
\newblock Articulated mesh animation from multi-view silhouettes.
\newblock \emph{{ACM} Trans. Graph.}, 27\penalty0 (3):\penalty0 97, 2008.

\bibitem[Vlasic et~al.(2009)Vlasic, Peers, Baran, Debevec, Popovic, Rusinkiewicz, and Matusik]{densecam_mesh_09}
Daniel Vlasic, Pieter Peers, Ilya Baran, Paul~E. Debevec, Jovan Popovic, Szymon Rusinkiewicz, and Wojciech Matusik.
\newblock Dynamic shape capture using multi-view photometric stereo.
\newblock \emph{{ACM} Trans. Graph.}, 28\penalty0 (5):\penalty0 174, 2009.

\bibitem[Wan et~al.(2023)Wan, Richardt, Bozic, Li, Rengarajan, Nam, Xiang, Li, Zhu, Ranjan, and Liao]{duplex}
Ziyu Wan, Christian Richardt, Aljaz Bozic, Chao Li, Vijay Rengarajan, Seonghyeon Nam, Xiaoyu Xiang, Tuotuo Li, Bo Zhu, Rakesh Ranjan, and Jing Liao.
\newblock Learning neural duplex radiance fields for real-time view synthesis.
\newblock In \emph{CVPR}, pages 8307--8316, 2023.

\bibitem[Wang et~al.(2021{\natexlab{a}})Wang, Liu, Liu, Theobalt, Komura, and Wang]{neus}
Peng Wang, Lingjie Liu, Yuan Liu, Christian Theobalt, Taku Komura, and Wenping Wang.
\newblock Neus: Learning neural implicit surfaces by volume rendering for multi-view reconstruction.
\newblock In \emph{NeurIPS}, pages 27171--27183, 2021{\natexlab{a}}.

\bibitem[Wang et~al.(2021{\natexlab{b}})Wang, Wang, Genova, Srinivasan, Zhou, Barron, Martin{-}Brualla, Snavely, and Funkhouser]{ibrnet}
Qianqian Wang, Zhicheng Wang, Kyle Genova, Pratul~P. Srinivasan, Howard Zhou, Jonathan~T. Barron, Ricardo Martin{-}Brualla, Noah Snavely, and Thomas~A. Funkhouser.
\newblock Ibrnet: Learning multi-view image-based rendering.
\newblock In \emph{CVPR}, pages 4690--4699, 2021{\natexlab{b}}.

\bibitem[Wang et~al.(2018)Wang, Liu, Zhu, Tao, Kautz, and Catanzaro]{highres}
Ting{-}Chun Wang, Ming{-}Yu Liu, Jun{-}Yan Zhu, Andrew Tao, Jan Kautz, and Bryan Catanzaro.
\newblock High-resolution image synthesis and semantic manipulation with conditional gans.
\newblock In \emph{CVPR}, pages 8798--8807, 2018.

\bibitem[Wang et~al.(2004)Wang, Bovik, Sheikh, and Simoncelli]{ssim}
Zhou Wang, Alan~C. Bovik, Hamid~R. Sheikh, and Eero~P. Simoncelli.
\newblock Image quality assessment: from error visibility to structural similarity.
\newblock \emph{{IEEE} Trans. Image Process.}, 13\penalty0 (4):\penalty0 600--612, 2004.

\bibitem[Wang et~al.(2023)Wang, Shen, Nimier-David, Sharp, Gao, Keller, Fidler, M\"uller, and Gojcic]{adashell}
Zian Wang, Tianchang Shen, Merlin Nimier-David, Nicholas Sharp, Jun Gao, Alexander Keller, Sanja Fidler, Thomas M\"uller, and Zan Gojcic.
\newblock Adaptive shells for efficient neural radiance field rendering.
\newblock \emph{ACM Trans. Graph.}, 42\penalty0 (6), 2023.

\bibitem[Weng et~al.(2022)Weng, Curless, Srinivasan, Barron, and Kemelmacher-Shlizerman]{humannerf}
Chung-Yi Weng, Brian Curless, Pratul~P. Srinivasan, Jonathan~T. Barron, and Ira Kemelmacher-Shlizerman.
\newblock Human{N}e{RF}: Free-viewpoint rendering of moving people from monocular video.
\newblock In \emph{CVPR}, pages 16210--16220, 2022.

\bibitem[Wu et~al.(2024)Wu, Mildenhall, Henzler, Park, Gao, Watson, Srinivasan, Verbin, Barron, Poole, et~al.]{reconfusion}
Rundi Wu, Ben Mildenhall, Philipp Henzler, Keunhong Park, Ruiqi Gao, Daniel Watson, Pratul~P Srinivasan, Dor Verbin, Jonathan~T Barron, Ben Poole, et~al.
\newblock Reconfusion: {3D} reconstruction with diffusion priors.
\newblock In \emph{CVPR}, pages 21551--21561, 2024.

\bibitem[Yu et~al.(2021{\natexlab{a}})Yu, Li, Tancik, Li, Ng, and Kanazawa]{octree}
Alex Yu, Ruilong Li, Matthew Tancik, Hao Li, Ren Ng, and Angjoo Kanazawa.
\newblock Plenoctrees for real-time rendering of neural radiance fields.
\newblock In \emph{ICCV}, pages 5732--5741, 2021{\natexlab{a}}.

\bibitem[Yu et~al.(2021{\natexlab{b}})Yu, Ye, Tancik, and Kanazawa]{pixelnerf}
Alex Yu, Vickie Ye, Matthew Tancik, and Angjoo Kanazawa.
\newblock pixelnerf: Neural radiance fields from one or few images.
\newblock In \emph{CVPR}, pages 4578--4587, 2021{\natexlab{b}}.

\bibitem[Yu et~al.(2020)Yu, Zhao, Zheng, Guo, Dai, Li, Pons{-}Moll, and Liu]{depth_single_20}
Tao Yu, Jianhui Zhao, Zerong Zheng, Kaiwen Guo, Qionghai Dai, Hao Li, Gerard Pons{-}Moll, and Yebin Liu.
\newblock Doublefusion: Real-time capture of human performances with inner body shapes from a single depth sensor.
\newblock \emph{{IEEE} Trans. Pattern Anal. Mach. Intell.}, 42\penalty0 (10):\penalty0 2523--2539, 2020.

\bibitem[Yu et~al.(2021{\natexlab{c}})Yu, Zheng, Guo, Liu, Dai, and Liu]{thu2}
Tao Yu, Zerong Zheng, Kaiwen Guo, Pengpeng Liu, Qionghai Dai, and Yebin Liu.
\newblock Function4d: Real-time human volumetric capture from very sparse consumer {RGBD} sensors.
\newblock In \emph{CVPR}, pages 5746--5756, 2021{\natexlab{c}}.

\bibitem[Zhang et~al.(2020)Zhang, Riegler, Snavely, and Koltun]{nerfpp}
Kai Zhang, Gernot Riegler, Noah Snavely, and Vladlen Koltun.
\newblock Nerf++: Analyzing and improving neural radiance fields.
\newblock \emph{CoRR}, abs/2010.07492, 2020.

\bibitem[Zhang et~al.(2018)Zhang, Isola, Efros, Shechtman, and Wang]{lpips}
Richard Zhang, Phillip Isola, Alexei~A. Efros, Eli Shechtman, and Oliver Wang.
\newblock The unreasonable effectiveness of deep features as a perceptual metric.
\newblock In \emph{CVPR}, pages 586--595, 2018.

\bibitem[Zheng et~al.(2024)Zheng, Zhou, Shao, Liu, Zhang, Nie, and Liu]{gps}
Shunyuan Zheng, Boyao Zhou, Ruizhi Shao, Boning Liu, Shengping Zhang, Liqiang Nie, and Yebin Liu.
\newblock Gps-gaussian: Generalizable pixel-wise 3d gaussian splatting for real-time human novel view synthesis.
\newblock In \emph{CVPR}, pages 19680--19690, 2024.

\bibitem[Zins et~al.(2023)Zins, Xu, Boyer, Wuhrer, and Tung]{srdf}
Pierre Zins, Yuanlu Xu, Edmond Boyer, Stefanie Wuhrer, and Tony Tung.
\newblock Multi-view reconstruction using signed ray distance functions {(SRDF)}.
\newblock In \emph{CVPR}, pages 16696--16706, 2023.

\end{thebibliography}
}

\ifarxiv \clearpage \appendix \section*{Appendix}
\label{sec:appendix_section}
\renewcommand{\thesubsection}{\Alph{subsection}}
\setcounter{subsection}{0}
In this appendix, we provide additional experimental results in Appendix \ref{sec:exp_add}, and additional implementation details in Appendix \ref{sec:impl_add}.
\subsection{Additional Experimental Results} \label{sec:exp_add}
We show additional quantitative and qualitative results to provide a comprehensive evaluation.
Table~\ref{tab:ft_sub} shows the quantitative results for each subject fine-tuned with images of target subjects.
Fig.~\ref{fig:qual_add} shows additional qualitative comparisons with the state-of-the-art methods.
Fig.~\ref{fig:abl_add} shows additional qualitative results for the ablation study on handling occluded regions.

\subsection{Additional Implementation Details} \label{sec:impl_add}
We first describe implementation details for method components described in Sec.~\ref{sec:geom}.
We use the first three operators of \textit{EfficientNetV2-S}~\cite{efficientnetv2} with an additional residual connection for our feature extraction CNN \knmrA{$\mathbf{E}$}. Please refer to Table~\ref{tab:feat_cnn} for network details \knmrA{for an} input image with size \knmrA{of} \(512\times512\).
\wA{
When sampling using the boundary meshes,
the self-intersections on the dilated SMPL\knmrA{-X} mesh do not \knmrA{matter} since we use either the first or the last intersection point.
The self-intersections on the eroded mesh, however, may cause missing geometries, especially for head, hands, and feet.
\knmrA{This is because the erosion process causes inward-facing surfaces in complicated body parts.}
Namely, some geometries are ``inside-out'', causing an
expansion on part of the eroded mesh and missing geometries.
Thus, we remove those body parts from the eroded mesh. Other body parts, i.e., torso, arms and legs have rather simple geometries and do not suffer the problem described above.
}
We implement the 3D CNN \knmrA{$\mathbf{G}$} following SparseConvNet~\cite{sparseconvnet} using \textit{TorchSparse} library~\cite{torchsparse} for efficient calculation.

For implementation details in Sec.~\ref{sec:render}, the architecture for \knmrA{$\mathbf{U}$} is based on the 2D neural rendering network described in GM-NeRF~\cite{gmnerf}. Please refer to Table~\ref{tab:occ_cnn} for details of our CNN \knmrA{$\mathbf{O}$} that predicts the occlusion map.
Finally, the architecture for our discriminator used for adversarial loss is based on the discriminator in LaMa~\cite{lama}.

\begin{figure*}[htb]
  \centering
{
  \setlength{\abovecaptionskip}{0pt}   
  \includegraphics[width=0.98\linewidth]{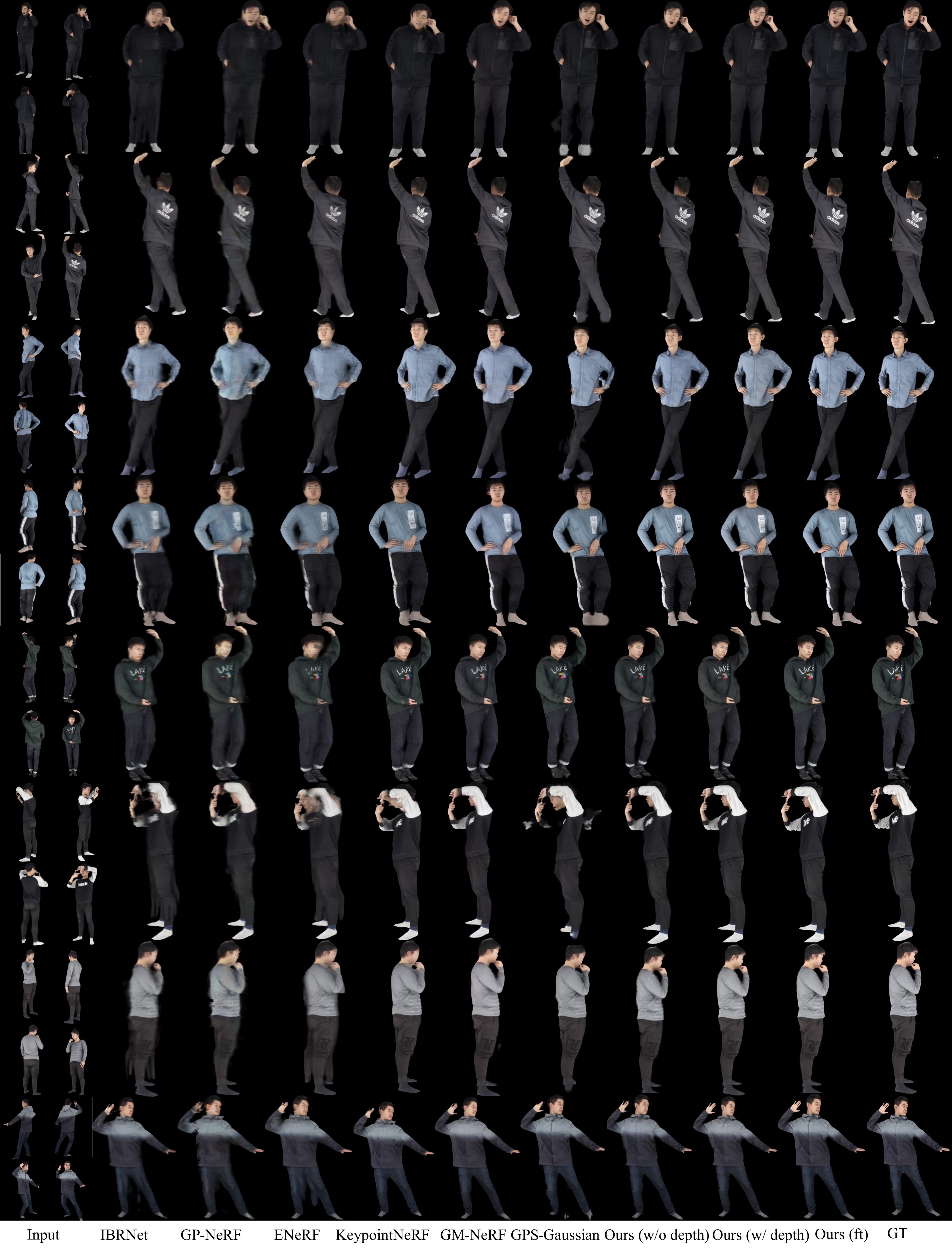} 
   \caption{Additional qualitative evaluation results on the THuman2.0 dataset compared with the state-of-the-art methods. The experimental settings are the same as in Fig.~\ref{fig:eval_qual}.}
    \label{fig:qual_add}
}
\end{figure*}
\begin{figure*}[htb]
  \centering
  \includegraphics[width=0.95\linewidth]{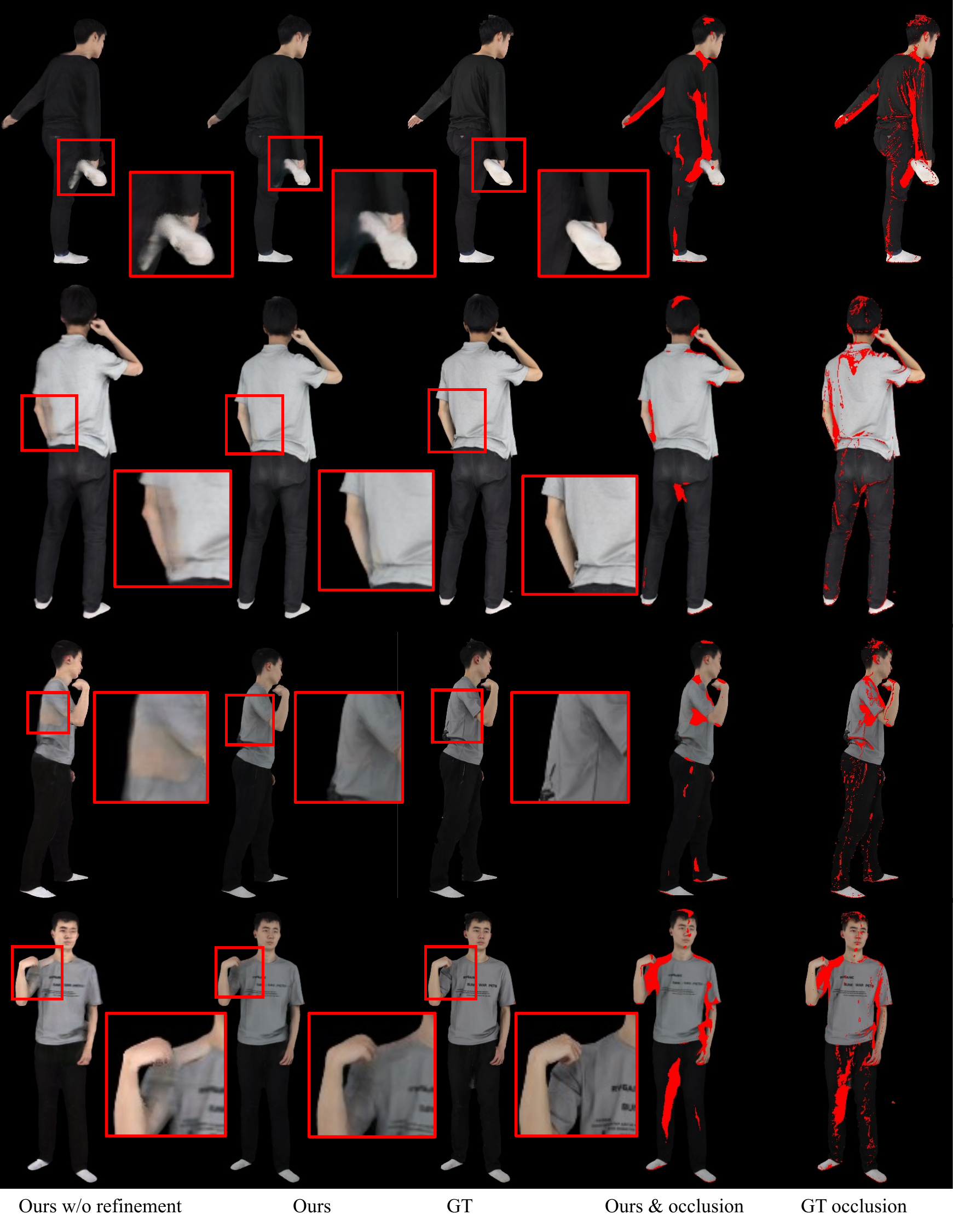}
  \caption{Additional ablation study results on handling occluded regions. The experimental settings are the same as in Fig.~\ref{fig:ablation2}.}
  \label{fig:abl_add}
\end{figure*}

\begin{table*}[ht]
  \centering
  \renewcommand{\arraystretch}{0.8} 
  \caption{Network details of the feature extraction CNN \(\mathbf{E}\) for an input image with size \(512\times512\times3\). Upsample\(\times2\) denotes the bi-linear upsampling operator with a scale factor of 2. ``\textbf{4}+\textbf{5}'' denotes the sum of the output feature maps with ID 4 and 5.}
  \label{tab:feat_cnn}
  \begin{tabular}{l|c|c|c|c|c}
      \toprule
      Operator & Kernel size & Stride & Input ID, Size (HWC) & Output ID, Size (HWC) & \#Layers \\
      \midrule
      Conv2D & 3 & 2 & \textbf{0}, \(512\times512\times3\) & \textbf{1}, \(256\times256\times24\) & 1 \\
      Fused\_MBConv~\cite{efficientnetv2} & 3 & 1 & \textbf{1}, \(256\times256\times24\) & \textbf{2}, \(256\times256\times24\) & 2 \\
      Fused\_MBConv~\cite{efficientnetv2} & 3 & 2 & \textbf{2}, \(256\times256\times24\) & \textbf{3}, \(128\times128\times48\) & 4 \\
      Upsample\(\times2\) & - & - & \textbf{3}, \(128\times128\times48\) & \textbf{4}, \(256\times256\times48\) & 1 \\
      Conv2D & 1 & 1 & \textbf{2}, \(256\times256\times24\) & \textbf{5}, \(256\times256\times48\) & 1 \\
      Conv2D & 3 & 1 & \textbf{4}+\textbf{5}, \(256\times256\times48\) & \textbf{6}, \(256\times256\times32\) & 1 \\
      \bottomrule
  \end{tabular}
\end{table*}
\begin{table*}[ht]
  \centering
  \renewcommand{\arraystretch}{0.8} 
  \caption{Network details of the feature extraction CNN \(\mathbf{O}\) for an input feature with size \(256\times256\times16\). ``ResBlock'' denotes the residual block with two convolution-batch-normalization-ReLU layers and a skip connection.}
  \label{tab:occ_cnn}
  \begin{tabular}{l|c|c|c|c|c}
      \toprule
      Operator & Kernel size & Stride & Input ID, Size (HWC) & Output ID, Size (HWC) & \#Layers \\
      \midrule
      Conv2D & 3 & 1 & \textbf{0}, \(256\times256\times16\) & \textbf{1}, \(256\times256\times96\) & 1 \\
      ResBlock & 3 & 1 & \textbf{1}, \(256\times256\times96\) & \textbf{2}, \(256\times256\times96\) & 2 \\
      TransposedConv & 3 & 2 & \textbf{2}, \(256\times256\times96\) & \textbf{3}, \(512\times512\times96\) & 1 \\
      Conv2D+Sigmoid & 1 & 1 & \textbf{3}, \(512\times512\times96\) & \textbf{4}, \(512\times512\times1\) & 1 \\
      \bottomrule
  \end{tabular}
\end{table*}
\begin{table*}
  \small
  \centering
  \setlength{\tabcolsep}{1pt} 
  \renewcommand{\arraystretch}{1.1} 
  \caption{Quantitative results for each subject fine-tuned with images of target subjects on the THuman2.0 dataset.} \label{tab:ft_sub}
  \begin{tabular}{|c|c|c|c||c|c|c|c||c|c|c|c||c|c|c|c|} \hline
  Subject & PSNR & SSIM & LPIPS & Subject & PSNR & SSIM & LPIPS & Subject & PSNR & SSIM & LPIPS & Subject & PSNR & SSIM & LPIPS \\ \hline
  401 & 31.95 & 0.989 & 0.018 & 402 & 34.06 & 0.989 & 0.023 & 403 & 23.69 & 0.933 & 0.073 & 404 & 39.03 & 0.991 & 0.015 \\ \hline
  405 & 41.01 & 0.994 & 0.013 & 406 & 34.99 & 0.993 & 0.016 & 407 & 33.93 & 0.990 & 0.017 & 408 & 34.08 & 0.979 & 0.036 \\ \hline
  409 & 39.37 & 0.995 & 0.012 & 410 & 37.46 & 0.993 & 0.015 & 411 & 34.83 & 0.988 & 0.020 & 412 & 38.14 & 0.995 & 0.011 \\ \hline
  413 & 35.96 & 0.993 & 0.016 & 414 & 35.11 & 0.992 & 0.016 & 415 & 36.57 & 0.992 & 0.016 & 416 & 37.97 & 0.991 & 0.015 \\ \hline
  417 & 35.46 & 0.992 & 0.015 & 418 & 35.43 & 0.992 & 0.017 & 419 & 38.18 & 0.993 & 0.016 & 420 & 38.35 & 0.993 & 0.015 \\ \hline
  421 & 37.01 & 0.993 & 0.016 & 422 & 41.91 & 0.994 & 0.014 & 423 & 39.16 & 0.994 & 0.013 & 424 & 39.44 & 0.994 & 0.012 \\ \hline
  425 & 38.00 & 0.993 & 0.014 & 426 & 38.65 & 0.991 & 0.015 & 427 & 39.38 & 0.995 & 0.013 & 428 & 37.81 & 0.993 & 0.017 \\ \hline
  429 & 37.80 & 0.994 & 0.013 & 430 & 38.06 & 0.992 & 0.015 & 431 & 39.31 & 0.994 & 0.013 & 432 & 38.19 & 0.995 & 0.011 \\ \hline
  433 & 38.11 & 0.995 & 0.011 & 434 & 37.84 & 0.991 & 0.015 & 435 & 38.14 & 0.985 & 0.028 & 436 & 33.34 & 0.960 & 0.054 \\ \hline
  437 & 39.04 & 0.992 & 0.016 & 438 & 35.75 & 0.990 & 0.020 & 439 & 39.31 & 0.995 & 0.014 & 440 & 35.65 & 0.994 & 0.016 \\ \hline
  441 & 38.44 & 0.994 & 0.015 & 442 & 37.04 & 0.989 & 0.020 & 443 & 32.36 & 0.984 & 0.024 & 444 & 32.92 & 0.988 & 0.021 \\ \hline
  445 & 37.72 & 0.990 & 0.018 & 446 & 41.52 & 0.996 & 0.010 & 447 & 38.55 & 0.993 & 0.017 & 448 & 39.63 & 0.996 & 0.013 \\ \hline
  449 & 40.87 & 0.994 & 0.013 & 450 & 37.53 & 0.994 & 0.013 & 451 & 34.97 & 0.993 & 0.012 & 452 & 36.64 & 0.993 & 0.013 \\ \hline
  453 & 35.10 & 0.992 & 0.014 & 454 & 35.82 & 0.994 & 0.012 & 455 & 33.22 & 0.988 & 0.018 & 456 & 33.51 & 0.989 & 0.018 \\ \hline
  457 & 36.45 & 0.993 & 0.014 & 458 & 37.40 & 0.994 & 0.014 & 459 & 36.46 & 0.993 & 0.016 & 460 & 35.17 & 0.992 & 0.017 \\ \hline
  461 & 35.26 & 0.992 & 0.014 & 462 & 34.88 & 0.991 & 0.014 & 463 & 35.01 & 0.989 & 0.018 & 464 & 40.25 & 0.993 & 0.015 \\ \hline
  465 & 39.37 & 0.988 & 0.024 & 466 & 38.59 & 0.991 & 0.016 & 467 & 36.11 & 0.986 & 0.026 & 468 & 38.46 & 0.995 & 0.011 \\ \hline
  469 & 37.13 & 0.994 & 0.013 & 470 & 35.93 & 0.994 & 0.014 & 471 & 39.59 & 0.995 & 0.011 & 472 & 38.95 & 0.994 & 0.013 \\ \hline
  473 & 38.80 & 0.994 & 0.013 & 474 & 38.83 & 0.992 & 0.014 & 475 & 38.70 & 0.992 & 0.016 & 476 & 38.03 & 0.993 & 0.013 \\ \hline
  477 & 36.41 & 0.992 & 0.014 & 478 & 41.85 & 0.995 & 0.010 & 479 & 40.41 & 0.995 & 0.014 & 480 & 40.13 & 0.994 & 0.013 \\ \hline
  481 & 31.96 & 0.979 & 0.036 & 482 & 35.86 & 0.992 & 0.015 & 483 & 38.65 & 0.994 & 0.013 & 484 & 30.40 & 0.977 & 0.033 \\ \hline
  485 & 35.59 & 0.992 & 0.016 & 486 & 36.44 & 0.992 & 0.015 & 487 & 36.86 & 0.993 & 0.017 & 488 & 31.89 & 0.980 & 0.030 \\ \hline
  489 & 37.42 & 0.995 & 0.011 & 490 & 34.79 & 0.989 & 0.016 & 491 & 36.98 & 0.993 & 0.015 & 492 & 36.96 & 0.994 & 0.012 \\ \hline
  493 & 35.97 & 0.991 & 0.018 & 494 & 28.08 & 0.973 & 0.037 & 495 & 35.49 & 0.994 & 0.013 & 496 & 34.55 & 0.991 & 0.015 \\ \hline
  497 & 32.79 & 0.983 & 0.029 & 498 & 36.59 & 0.993 & 0.013 & 499 & 38.21 & 0.992 & 0.016 & 500 & 37.76 & 0.992 & 0.015 \\ \hline
  501 & 37.51 & 0.991 & 0.017 & 502 & 38.18 & 0.995 & 0.013 & 503 & 38.60 & 0.994 & 0.015 & 504 & 39.48 & 0.994 & 0.014 \\ \hline
  505 & 39.42 & 0.995 & 0.009 & 506 & 40.75 & 0.993 & 0.015 & 507 & 34.35 & 0.987 & 0.018 & 508 & 24.98 & 0.818 & 0.085 \\ \hline
  509 & 23.47 & 0.788 & 0.075 & 510 & 38.88 & 0.995 & 0.011 & 511 & 35.99 & 0.987 & 0.029 & 512 & 31.52 & 0.982 & 0.023 \\ \hline
  513 & 37.14 & 0.990 & 0.017 & 514 & 35.53 & 0.993 & 0.015 & 515 & 33.92 & 0.988 & 0.021 & 516 & 33.91 & 0.990 & 0.020 \\ \hline
  517 & 30.89 & 0.981 & 0.031 & 518 & 32.43 & 0.986 & 0.023 & 519 & 33.54 & 0.992 & 0.012 & 520 & 33.05 & 0.990 & 0.013 \\ \hline
  521 & 34.30 & 0.992 & 0.011 & 522 & 32.53 & 0.988 & 0.021 & 523 & 33.38 & 0.989 & 0.019 & 524 & 34.46 & 0.993 & 0.013 \\ \hline
  525 & 33.52 & 0.990 & 0.018 & Average & 36.29 & 0.988 & 0.018 &  &  &  &  &  &  &  &  \\ \hline
  \end{tabular}
\end{table*} \fi

\end{document}